\documentclass[preprint,12pt,authoryear]{elsarticle}

\usepackage{amssymb}
\usepackage{amsmath}
\usepackage{natbib}
\setcitestyle{numbers}
\usepackage{subfig}
\usepackage{url} 
\usepackage{cancel}

\journal{Applied Soft Computing}

\begin{document}

\begin{frontmatter}



\title{Jacobian-Enhanced Neural Networks} 


\author[ELSYS]{Steven H. Berguin} 

\affiliation[ELSYS]{organization={Sr Researcher Engineer at Georgia Tech Research Institute},
            city={Atlanta},
            state={GA},
            postcode={30332}, 
            country={USA}
}
            
\begin{abstract}
Jacobian-Enhanced Neural Networks (JENN) are densely connected multi-layer perceptrons, whose training process is modified to predict partial derivatives more accurately. Their main benefit is better accuracy with fewer training points compared to standard neural networks. These attributes are particularly desirable in the field of design optimization, where there is often the need to replace computationally expensive models with fast running approximations, known as surrogate models (or metamodels), to speed up solutions. In the special case of gradient-enhanced methods, there is the additional value proposition that partial derivatives are accurate, which yields improved accuracy for gradient-based optimization. This work develops the theory of Jacobian-Enhanced Neural Nets and applies it to an airfoil optimization problem in order to illustrate its benefits over the \textit{status quo}.  
\end{abstract}


\begin{highlights}
\item A neural network formulation capable of predicting partial derivatives more accurately with significantly fewer training data samples
\item Practical illustration through applied aerospace engineering example, taken to be aerodynamic shape optimization using CFD  
\item Complete mathematical derivation of the underlying theory for arbitrary deep network, including vectorized equations for computer programming, with open-source Python code implementation
\item Quantification of advantages \& limitations through canonical example, including analysis of noisy partials and performance on benchmark optimization problem
\item Convenient framework to handle missing partials, believed to be a common use-case of practical importance.
\end{highlights}


\end{frontmatter}



\section*{Nomenclature}

\begin{tabbing}
  XXX \= \kill
  $L$ \> number of layers in neural network \\
  $a$ \> activation output, $a = g(z)$ \\
  $b$ \> bias (parameter to be learned) \\
  $g$ \> activation function \\
  $m$ \> number of training examples \\
  $n$ \> number of nodes in a layer \\
  $n_x$ \> number of nodes in input layer, $n_x=n^{[0]}$ \\
  $n_y$ \> number of nodes in output layer, $n_y=n^{[L]}$ \\
  $w$ \> weight (parameter to be learned) \\
  $x$ \> training data input \\
  $y$ \> training data output \\
  $\hat{y}$ \> predicted output \\
  $z$ \> linear activation input \\
  \\
  \textit{Hyperparameters}\\
  $\alpha$ \> learning rate \\
  $\beta$ \> hyperparameter to priorities training data values \\
  $\gamma$ \> hyperparameter to priorities training data partials \\
  $\lambda$ \> hyperparameter used for regularization  \\
  \\
  \textit{Mathematical Symbols and Operators}\\
  $\mathcal{J}$ \> cost function \\
  $\mathcal{L}$ \> loss function \\
  $()_j\prime$ \> first derivative, $\partial () / \partial x_j$\\
  $()_j\prime\prime$ \> second derivative, $\partial^2 () / \partial x_j^2$ \\
  \\
  \textit{Superscripts/Subscripts}\\
  $i$ \> $i^\text{th}$ node in current layer, $1 \le i \le n^{[l]}$ \\
  $j$ \> $j^\text{th}$ input, $x_j$, $1 \le j \le n_x$  \\
  $k$ \> $k^\text{th}$ output, $y_k$, $1 \le k \le n_y$ \\
  $l$ \> $l^\text{th}$ layer of neural network, $1 \le l \le L$ \\
  $r$ \> $r^\text{th}$ node of {\it previous} layer, $1 \le r \le n^{[l-1]}$ \\
  $s$ \> $s^\text{th}$ node of {\it current} layer, $1 \le s \le n^{[l]}$ \\
  $t$ \> $t^\text{th}$ training example, $1 \le t \le m$ 
 \end{tabbing}

\section{Introduction}

Aerodynamic shape optimization is concerned with maximizing aerodynamic performance by changing the shape of a body subjected to specific flow conditions. The state of the art uses computational fluid dynamics and adjoint methods to achieve this goal~\cite{Jameson1988, Kenway2019}. These methods scale very well, enabling optimization problems with hundreds of decision variables to be solved as efficiently as problems with a handful. However, the cost of individual function calls remains an impediment: a single flow solution can take several hours depending on the use case. For this reason, there has been significant interest in accelerating numerical computations through machine learning in aerodynamic shape optimization~\cite{Li2022}, and the present work focuses on one method in particular: Jacobian-Enhanced Neural Networks (JENN).     

\subsection{Overview}

JENN are fully connected multi-layer perceptrons, whose training process has been modified to predict partial derivatives more accurately. This is accomplished by minimizing a modified version of the Least Squares Estimator (LSE) that accounts for Jacobian prediction error, where a Jacobian extends the concept of the gradient for functions of more than one output. The main benefit of Jacobian enhancement is better accuracy with fewer training points compared to standard fully connected neural nets. This paper develops the theory behind JENN and applies it to an airfoil optimization problem in order to illustrate its benefits over \textit{status quo} neural networks. It complements previous work on gradient-enhanced neural nets in three ways. First, it provides a concise and complete derivation of the theory, including equations in vectorized form, without relying on automatic differentiation. Second, it provides a deep-learning formulation that can handle any number of inputs, hidden layers, or outputs. Third, through judicious use of hyperparameters, it enables individual sample points to be prioritized differently in situations that call for it, the utility of which is illustrated in the application section. 

\subsection{Intended Use-Case} 

Even though JENN was developed with aerodynamic shape optimization in mind, it is a general method that applies to any engineering problem for which partial derivatives are defined.  It is intended for the field of computer aided design, where there is often a need to replace computationally expensive simulation models with fast running approximations, commonly referred to as surrogate models or metamodels. Since a surrogate emulates the original model with close approximation in a fraction of a second, it yields a speed benefit that enables orders of magnitude more function calls to be carried out quickly. This allows designers to explore a tradespace more fully, in order to better understand the impact of design variables on the performance of the system under design. While this is true of any surrogate modeling approach, JENN has an additional benefit that most surrogate modeling approaches lack: the ability to predict partial derivatives more accurately. This is a desirable property for one use case in particular: surrogate-based optimization. The field of engineering is rich in design applications of such a use case with entire book chapters dedicated to it~\cite{Keane2005, Forrester2011, Iuliano2016, Martins_Ning_2021}, review papers~\cite{Queipo2005, haridy2023review, su14073867, Wang2006}, and countless examples to be found, particularly in the field of aerospace engineering~\cite{Robinson2008, Giannakoglou2006, Bouhlel2020, Nagawkar2022}.  

\subsection{Limitations} 

First and foremost, gradient-enhanced methods require factors and responses to be smooth and continuous. Second, the value proposition of a surrogate is that the computational expense of generating training data to fit the model is much less than the computational expense of directly performing the analysis with the original model itself, which is not always true, especially when computing partials. Gradient-enhanced methods are therefore only beneficial if the cost of obtaining partials is not excessive in the first place or if the need for accuracy outweighs the cost of computing the partials; see Martins et al.~\cite{Martins2013a} for an excellent review of all existing methods for computing the derivatives of
computational models.   
\section{Previous Work}

A cursory literature research reveals that the field of gradient-enhanced learning is predominantly focused on Gaussian processes. Dedicated book chapters can be found about gradient-enhanced Kriging models~\cite{Forrester2011}, salient open-source libraries have been almost exclusively devoted to gradient-enhanced Bayesian methods~\cite{saves2024smt}, and even review papers about gradient-enhanced metamodels only mention neural networks in passing~\cite{Laurent2019}. However, early work by~\cite{Sellar1996} can be found which uses gradient-enhanced neural network approximations in the context of Concurrent Subspace Optimization (CSSO), a type of optimization architecture in the field of multi-disciplinary design optimization. The authors of that work conclude that gradient-enhanced neural networks are more accurate and require significantly less training data than their non-enhanced counterparts, which seems to be the general consensus in all surveyed work to follow. 

Follow-on research by \cite{Liu2000} further provides the mathematical derivation of the equations for neural network gradient-enhancement, but the formulation is not general; it is limited to shallow neural networks with only two layers, which is of limited utility. This shortcoming is overcome by \cite{Giannakoglou2006} who developed a deep neural network formulation with gradient enhancement and used it for aerodynamic design applications. All necessary equations to reproduce their algorithm are provided, but their formulation is limited to only one output and the vectorized form is missing for computer programming. 

More recent work by \cite{Bouhlel2020} takes a different approach, relying instead on established deep-learning frameworks to delegate back-propagation to third-party, automatic differentiation tools. Using TensorFlow~\cite{tensorflow2015}, gradient-enhanced neural networks are successfully used to accelerate airfoil shape optimization, culminating into \textit{webfoil}\footnote{\url{https://webfoil.engin.umich.edu/}}; a useful web App for fast airfoil shaping. Overall, those authors reaffirm the previous conclusion that gradient-enhanced techniques improve accuracy and reduce the amount of training data required for good fits. While deep learning frameworks have obvious advantages (\textit{e.g.} parallel computing), there is complementary research value in developing the equations for back-propagation from scratch, which is one of the objectives of this paper. Stand-alone equations are reproducible without third-party tools and convey helpful insight into the mathematics behind a specific algorithm. Backprop is a special case of automatic differentiation for scalar functions. 

The works surveyed so far have focused on prediction accuracy, but a keyword search for Jacobian and neural networks also points to the concept of Jacobian regularization \cite{hoffman2019robust} for classification. The key idea is to ensure robustness of models against input perturbations due to adversarial attacks (\textit{e.g.} a purposefully modified input image to avoid detection) by using Jacobian information to reduce sensitivity. However, robustness is outside the scope of the present work, which seeks to predict partials accurately for design applications given high quality inputs.  

Finally, the most recent work on gradient-enhanced neural networks focuses on gradient-enhanced Physics-Informed Neural Networks (PINN)~\cite{yu2022}, which uses the residuals of partial differential equations as the loss function for training neural nets. It is a deep learning framework for predicting systems governed by nonlinear partial differential equations, in which the neural network effectively learns to solve a system of equations rather than merely emulating a dataset. However, PINN falls outside the scope of the present work, which seeks to develop a general purpose algorithm to fit any smooth function regardless of what it represents, be it physics-informed or not. 
\section{Mathematical Derivation} 

\begin{figure}[!ht]
  \centering
  \includegraphics[width=1.75in]{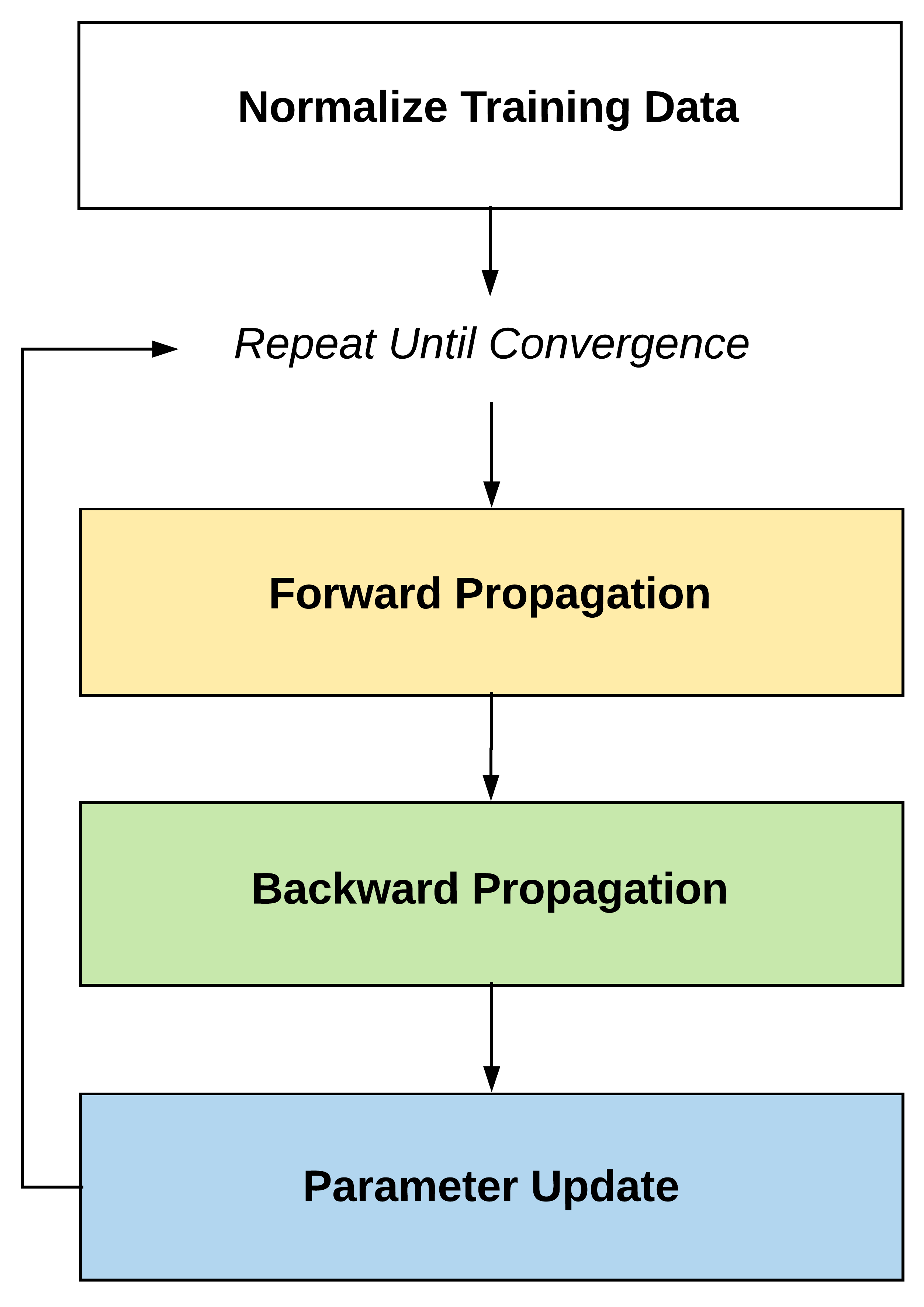}
  \caption{summary of neural net training process}
  \captionsetup{justification=centering}
  \label{fig:ANN}
\end{figure}

Jacobian-Enhanced Neural Networks (JENN) follow the same steps as standard Neural Networks (NN), shown in Fig.~\ref{fig:ANN}, except that the cost function used during parameter update is modified to account for partial derivatives, with the consequence that back propagation must be modified accordingly. This section provides the mathematical derivation.  

\subsection{Notation}

For clarity, let us adopt a similar notation to the one used by \cite{Ng2017} where superscript $l$ denotes the layer number, $t$ is the training example, and subscript $i$ indicates the node number in a layer. For instance, the activation of the $i^\text{th}$ node in layer $l$ evaluated at example $t$ would be denoted:   
\begin{align*}
a_i^{[l](t)} = g\left(z_i^{[l](t)}\right) 
\quad 
\begin{cases}
\forall~ i \in \{1, \dots, n^{[l]}\}  \\
\forall~ l \in \{1, \dots, L\}  \\
\forall~ t \in \{1, \dots, m\}  
\end{cases}
\end{align*}
To avoid clutter in the derivation to follow, it will be understood that subscripts $s$ and $r$ refer to quantities associated with the current and previous layer, respectively, evaluated at example $t$ without the need for superscripts. Concretely:
\begin{subequations}
\begin{alignat}{4}
a_s &\equiv a_{i=s}^{[l](t)} \quad && z_s &&\equiv z_{i=s}^{[l](t)} 
~ &&\forall~ s \in \{1, \dots, n^{[l]}\} \\
a_r &\equiv a_{i=r}^{[l-1](t)} \quad && z_r &&\equiv z_{i=r}^{[l-1](t)} 
~  &&\forall~ r \in \{1, \dots, n^{[l-1]}\} 
\end{alignat}
\end{subequations}
Derivatives w.r.t.input $x_j$ are denoted by the \textit{prime} symbol and subscript $j$, \textit{e.g.}
\begin{alignat*}{4}
a_{s_j}^{\prime} &\equiv \left(\dfrac{\partial a_s}{\partial x_j}\right)
\end{alignat*}
In turn, vector quantities will be defined in {\bf bold} font. For instance, the input and output layers associated with example $t$ would be given by: 
\begin{align*}
\boldsymbol{a}^{[1](t)} = \boldsymbol{x}^{(t)} = \left[ \begin{matrix} x_1^{(t)} \\ \vdots \\ x_{n_x}^{(t)}\end{matrix} \right] 
\quad 
\boldsymbol{y}^{(t)} = \left[ \begin{matrix} {y}_1^{(t)} \\ \vdots \\ {y}_{n_y}^{(t)}\end{matrix} \right] = \boldsymbol{a}^{[L](t)}
\end{align*}
%


\subsection{Normalize Training Data}

Normalizing the training data can greatly improve performance when either inputs or outputs differ by orders of magnitude. A simple way to improve numerical conditioning is to subtract the mean $\mu$ and divide by the standard deviation $\sigma$ of the data, provided the denominator is not close to zero. 
The normalized data then becomes: 
\begin{subequations}
\begin{align}
\boldsymbol{x}^{(t)} &:= \frac{\boldsymbol{x}^{(t)} - \boldsymbol{\mu}_{x}}{\boldsymbol{\sigma}_{x}} ~ \in ~ \mathbb{R}^{n_x} 
\\
\boldsymbol{y}^{(t)} &:= \frac{\boldsymbol{y}^{(t)} - \boldsymbol{\mu}_{y}}{\boldsymbol{\sigma}_{y}} ~ \in ~ \mathbb{R}^{n_y} 
\\
{y}_{k_j}^{\prime(t)} &:= \left(\frac{\partial y_k}{\partial x_j}\right)^{(t)} 
\times \frac{\sigma_{x_j}}{{\sigma}_{y_k}}
~ \in ~ \mathbb{R}
\end{align}
\end{subequations}

\subsection{Forward Propagation} 

Predictions are obtained by propagating information forward throughout the neural network in a recursive manner, starting with the input layer and ending with the output layer. For each layer (excluding input layer), activations are computed as:  
\begin{subequations} \label{forward_prop}
\begin{alignat}{3}
& a_s = g(z_s) 
& \qquad 
& z_s = w_{sr} a_r + b_s 
\\
& a_{s_j}^\prime = g^\prime(z_s) z_{s_j}^\prime
& \qquad 
& z_{s_j}^\prime = w_{sr} a_{r_j}^\prime 
\end{alignat}
\end{subequations}
where $\left\{w_{sr}, b_s\right\}$ are neural network parameters to be learned. For the hidden layers ($1 < l < L$), the activation function is taken to be the hyperbolic tangent\footnote{Other choices are possible, provided the function is smooth (\textit{i.e.} not ReLu)}: 
\begin{subequations}
\begin{align}
g(z) &= \frac{e^{z}-e^{-z}}{e^{z}+e^{-z}} 
\\
g^\prime(z) &\equiv \frac{\partial g}{\partial z} = 1 - g(z)^2 
\\
g^{\prime\prime}(z) &\equiv \frac{\partial^2 g}{\partial z^2} = -2g(z)g(z)^\prime
\end{align}
\end{subequations}
For the output layer ($l=L$), the activation function is linear: 
\begin{align}
g(z) = z \quad \Rightarrow \quad g^\prime(z) = 1 \quad \Rightarrow \quad  g^{\prime\prime}(z) = 0
\end{align}

\subsection{Parameter Update}

Neural network parameters are learned by solving an unconstrained optimization problem: 
\begin{align}
\theta^{*} = \arg \underset{\theta}{\min} ~ \mathcal{J(\theta)}
\quad \forall \quad 
\theta \in \left\{w_{sr}, b_s\right\} 
\end{align}
where it should be understood that $\left\{w_{sr}, b_s\right\}$ is short hand for all parameters of the neural network. The cost function $\mathcal{J}$ is given by the following expression, where $\lambda$ is a hyperparameter to be tuned which controls regularization:  
\begin{align} \label{cost}
\mathcal{J(\theta)} &= \frac{1}{m} \sum_{t=1}^m  \mathcal{L}^{(t)} + \frac{\lambda}{2m} \sum_{l=2}^L \sum_{s=1}^{n^{[l]}} \sum_{r=1}^{n^{[l-1]}} w_{sr}^2 
\end{align}
Parameters are updated iteratively using \textit{gradient-descent}: 
\begin{align}
\label{GD}
\theta &:= \theta - \alpha \frac{\partial \mathcal{J}}{\partial \theta} 
\end{align}
where the hyperparameter $\alpha$ is the learning rate. In practice, some improved version of \textit{gradient-descent} would be used instead, such as ADAM~\cite{kingma2017adam}, but the former is easier to explain without loss of generality. The loss function $\mathcal{L}$ is taken to be a modified version of the Least Squares Estimator (LSE), augmented with an extra term to account for partial derivatives: 
\begin{align} \label{eq:loss}
\mathcal{L}^{(t)}
= 
\frac{1}{2}
&\sum_{s=1}^{n_y}
\beta_s^{(t)}
\left(a_s^{[L](t)} - y_s^{(t)}\right)^2 
+
\frac{1}{2}
\sum_{s=1}^{n_y}
\sum_{j=1}^{n_x} 
\gamma_{s_j}^{(t)}
\left( a_{s_j}^{\prime[L](t)} - y_{s_j}^{\prime(t)} \right)^2
\end{align}
Note that two more hyperparameters have been introduced: $\beta^{(t)}$ and $\gamma^{(t)}$. They act as weights that can be used to prioritize subsets of the training data in situations that call for it (example later). By default, $\beta_s^{(t)} = \gamma_{s_j}^{(t)} = 1$, meaning that all data is prioritized equally. 

\subsection{Backward Propagation}

Forward propagation seeks to predict the gradient of outputs $\boldsymbol{y}$ with respect to 
 (abbreviated w.r.t.) inputs $\boldsymbol{x}$. By contrast, back-propagation computes the gradient of the cost function $\mathcal{J}$ w.r.t. neural net parameters $\theta$:
\begin{align} \label{cost-derivative}
\frac{\partial \mathcal{J}}{\partial \theta} 
&= 
\frac{1}{m} \sum_{t=1}^m  \frac{\partial \mathcal{L}}{\partial \theta}^{(t)}
+ 
\frac{\lambda}{2m} \sum_{l=2}^L \sum_{s=1}^{n^{[l]}} \sum_{r=1}^{n^{[l-1]}} \frac{\partial }{\partial \theta} \left( w_{sr}^2 \right)
\end{align}

\subsubsection{Loss Function Derivatives w.r.t. Neural Net Parameters}

In a first step, assume the quantities ${\partial \mathcal{L}}/{\partial a_s}$ and ${\partial \mathcal{L}}/{\partial a^\prime_{s_j}}$ are known and consider the gradient of the loss function $\mathcal{L}$ w.r.t.the parameters $\theta$ from layer $l$. Applying the chain rule to Eq.~\ref{eq:loss} yields: 
\begin{align} 
\label{eq:dtheta}
\frac{\partial \mathcal{L}}{\partial \theta}
=
\frac{\partial \mathcal{L}}{\partial a_s}
\frac{\partial a_s}{\partial z_s}
\frac{\partial z_s}{\partial \theta}
+
\sum_{j=1}^{n_x}
\frac{\partial \mathcal{L}}{\partial a^\prime_{s_j}}
\left(
\frac{\partial a^\prime_{s_j}}{\partial z_s}
\frac{\partial z_s}{\partial \theta}
+
\frac{\partial a^\prime_{s_j}}{\partial z^\prime_{s_j}}
\frac{\partial z^\prime_{s_j}}{\partial \theta}
\right)
\end{align}
where the superscript $t$ has been dropped for notational convenience, with the understanding that all equations to follow now apply to example $t$. The various partial derivatives in Eq.~\ref{eq:dtheta} follow from Eq.~\ref{forward_prop}: 
\begin{alignat}{3} 
& \frac{\partial a_s}{\partial z_s} = g^\prime(z_s) 
\quad 
&&\frac{\partial a_{s_j}^\prime}{\partial z_s} = g^{\prime\prime}(z_s)z_{s_j}^\prime
\quad
&&\frac{\partial a_{s_j}^\prime}{\partial z_{s_j}^\prime} = g^{\prime}(z_s) 
\\
& \theta = w_{sr}: 
\quad 
&&\frac{\partial z_s}{\partial \theta} = a_r
\quad 
&&\frac{\partial z^\prime_{s_j}}{\partial \theta} = a_{r_j}^\prime
\\
& \theta = b_s: 
\quad 
&&\frac{\partial z_s}{\partial \theta} = 1
\quad 
&&\frac{\partial z^\prime_{s_j}}{\partial \theta} = 0
\end{alignat}
Substituting these expressions into Eq.~\ref{eq:dtheta} yields: 
\begin{subequations}\label{eq:dParams}
\begin{align}
\frac{\partial \mathcal{L}}{\partial w_{sr}} \label{eq:dW}
&=
\frac{\partial \mathcal{L}}{\partial a_s}
g^\prime(z_s) 
a_r 
+ \sum_{j=1}^{n_x}
\frac{\partial \mathcal{L}}{\partial a^\prime_{s_j}}
\Bigl(
g^{\prime\prime}(z_s)z_{s_j}^\prime
a_r
+
g^{\prime}(z_s)
a_{r_j}^\prime
\Bigr) 
\\
\frac{\partial \mathcal{L}}{\partial b_s} \label{eq:db}
&=
\frac{\partial \mathcal{L}}{\partial a_s}
g^\prime(z_s) 
+
\sum_{j=1}^{n_x}
\frac{\partial \mathcal{L}}{\partial a^\prime_{s_j}}
g^{\prime\prime}(z_s)z_{s_j}^\prime
\end{align}
\end{subequations}

\subsubsection{Loss Function Derivatives w.r.t. to Neural Net Activations}

All quantities in Eq.~\ref{eq:dParams} are known, except for ${\partial \mathcal{L}}/{\partial a_s}$ and ${\partial \mathcal{L}}/{\partial a_{s_j}^{\prime}} $ which must be obtained recursively by propagating information backwards from one layer to the next, starting with output layer $L$. From Eq.~\ref{eq:loss}, it follows that:  
\begin{subequations}
\label{eq:last-layer}
\begin{align}
\frac{\partial \mathcal{L}}{\partial a_s}^{[L]} &= \beta_s \left( a_s - y_s\right) 
\\
\frac{\partial \mathcal{L}}{\partial a_{s_j}^{\prime}}^{[L]} &= \gamma_{s_j} \left( a_{s_j}^{\prime} - y_s^{\prime}\right)
\end{align}
\end{subequations}
The derivatives for the next layer are obtained through the chain rule. According to Eq.~\ref{eq:loss}, the loss function $\mathcal{L}$ depends on $n^{[l]}$ activations $a_s$ and $n_x\times n^{[l]}$ associated partials $a_{s_j}^\prime$, which are themselves functions of $a_r$ and $a_{r_j}^\prime$. Hence, according to the chain rule, previous layer derivatives w.r.t. to current layer activations are:
\begin{subequations}
\label{eq:dActivations}
\begin{alignat}{2}
\label{eq:dA}
\frac{\partial \mathcal{L}}{\partial a_r}^{[l]}
&=
\sum_{s=1}^{n^{[l]}}
\left[
\frac{\partial \mathcal{L}}{\partial a_s}
\frac{\partial a_s}{\partial a_r}
+
\sum_{j=1}^{n_x}
\frac{\partial \mathcal{L}}{\partial a_{s_j}^\prime}
\frac{\partial a_{s_j}^\prime}{\partial a_r}
\right]
\quad &\forall ~ l = L, \dots, 2 
\\
\label{eq:dA_prime}
\frac{\partial \mathcal{L}}{\partial a^\prime_{r_j}}^{[l]}
&=
\sum_{s=1}^{n^{[l]}}
\left[
\frac{\partial \mathcal{L}}{\partial a_s}
\frac{\partial a_s}{\partial a^\prime_{r_j}}
+
\sum_{j=1}^{n_x}
\frac{\partial \mathcal{L}}{\partial a_{s_j}^\prime}
\frac{\partial a_{s_j}^\prime}{\partial a^\prime_{r_j}}
\right]
\quad &\forall ~ l = L, \dots, 2 
\end{alignat}
\end{subequations}
Further applying the chain rule to the intermediate quantities and using Eq.~\ref{forward_prop}, one finds: 
\begin{subequations}
\begin{align}
\frac{\partial a_s}{\partial a_r} 
&= 
\frac{\partial a_s}{\partial z_s} \frac{\partial z_s}{\partial a_r} 
+
\frac{\partial a_s}{\partial z_{s_j}^\prime} 
\cancelto{0}{\frac{\partial z_{s_j}^\prime}{\partial a_r}}
=
g^\prime(z_s) w_{sr}
\\
\frac{\partial a_s}{\partial a^\prime_{r_j}} &= 
\cancelto{0}{\frac{\partial a_s}{\partial z_s}} \frac{\partial z_s}{a^\prime_{r_j}}
+
\cancelto{0}{\frac{\partial a_s}{\partial z_{s_j}^\prime}}
\frac{\partial z_{s_j}^\prime}{a^\prime_{r_j}}
= 0
\\
\frac{\partial a_{s_j}^\prime}{\partial a_r} &= 
\frac{\partial a_{s_j}^\prime}{\partial z_s} \frac{\partial z_s}{\partial a_r} 
+
\frac{\partial a_{s_j}^\prime}{\partial z_{s_j}^\prime} 
\cancelto{0}{\frac{\partial z_{s_j}^\prime}{\partial a_r}}
= g^{\prime\prime}(z_s) z_{s_j}^\prime w_{sr}
\\
\frac{\partial a_{s_j}^\prime}{\partial a^\prime_{r_j}} &=
\frac{\partial a_{s_j}^\prime}{\partial z_s} \cancelto{0}{\frac{\partial z_s}{\partial a^\prime_{r_j}} }
+
\frac{\partial a_{s_j}^\prime}{\partial z_{s_j}^\prime} \frac{\partial z_{s_j}^\prime}{\partial a^\prime_{r_j}} 
=
g^\prime (z_s) w_{sr}
\end{align}
\end{subequations}
All necessary equations to compute the derivatives of the loss function w.r.t. parameters in any layer are now known. Derivatives are obtained by working our way back recursively across layers, repeatedly applying Eq.~\ref{eq:dActivations} after passing the solution in the current layer to the preceding layer as follows:   
\begin{subequations}
\begin{align}
\frac{\partial \mathcal{L}}{\partial a_s}^{[l-1]} &= \frac{\partial \mathcal{L}}{\partial a_r}^{[l]} 
\\
\frac{\partial \mathcal{L}}{\partial a^\prime_{s_j}}^{[l-1]} &= \frac{\partial \mathcal{L}}{\partial a^\prime_{r_j}}^{[l]}
\end{align}
\end{subequations}

\subsubsection{Cost Function Derivatives w.r.t. Parameters}

Putting it all together, the derivative of the cost function w.r.t. neural network parameters are obtained by substituting Eq.~\ref{eq:dParams} into Eq.~\ref{cost-derivative}, where the subscripts $s$ and $r$ are referenced to layer $l$:
\begin{subequations}
\label{eq:cost-derivatives}
\begin{alignat}{2} 
\frac{\partial \mathcal{J}}{\partial w_{sr}}
&= 
\frac{1}{m}
\sum_{t=1}^{m}
\frac{\partial \mathcal{L}^{(t)}}{\partial w_{sr}}
+
\lambda w_{sr} 
&\quad\forall~1<l\le L
\\
\frac{\partial \mathcal{J}}{\partial b_s}
&=
\frac{1}{m}
\sum_{t=1}^{m}
\frac{\partial \mathcal{L}^{(t)}}{\partial b_s}
&\quad\forall~1<l\le L
\end{alignat}
\end{subequations}
A summary of key equations is available in \ref{app:vectorized-equations} in vectorized form for computer programming, which have already been implemented in an open-source library called JENN developed for this work.
\section{Implementation}

An open-source Python library that implements JENN can be found online\footnote{\url{https://pypi.org/project/jenn/}}. There exist many excellent deep learning frameworks, such as TensorFlow~\cite{tensorflow2015} and PyTorch~\cite{PyTorch}, as well as many all-purpose machine learning libraries such as scikit-learn~\cite{sklearn2011}. However,  gradient-enhancement is not inherently part of those popular frameworks, let alone the availability of prepared functions to predict partials after training is complete, which requires non-trivial effort to be implemented in those tools. The present library seeks to close that gap. It is intended for non-practitioners, who have access to partials and are seeking a minimal-effort Application Programming Interface (API) with low-barrier to entry to take advantage of them. JENN focuses exclusively on neural networks and  serves as an upstream dependency to the popular Surrogate Modeling Toolbox (SMT)~\cite{saves2024smt} specializing in gradient-enhanced methods in general. 

\subsection{Verification and Validation}

\begin{figure*}[!t]
\centering
\subfloat[$y=sin(x)$]{
    \includegraphics[width=1.75in]{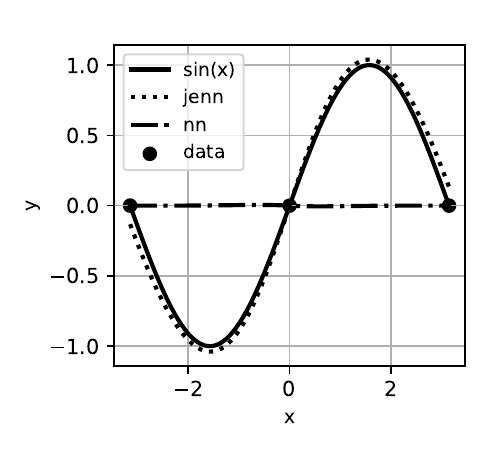}%
    \label{subfig:validation:sinx}
}
\subfloat[$y=x sin(x)$]{
    \includegraphics[width=1.75in]{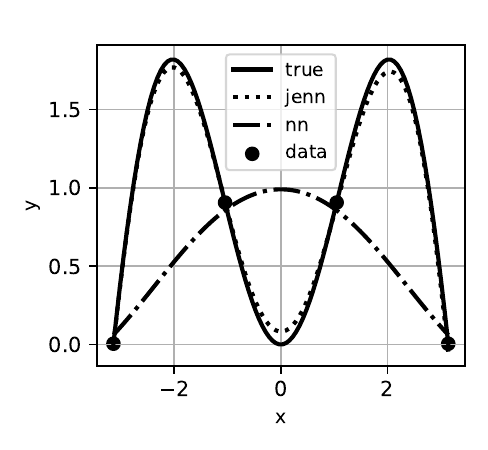}%
    \label{subfig:validation:xsinx}
}
\\
\subfloat[$y=\sum_{i=1}^2\left(x_i^2 - 10 \cos(2\pi x_i)+10\right)$]{
    \includegraphics[height=1.5in]{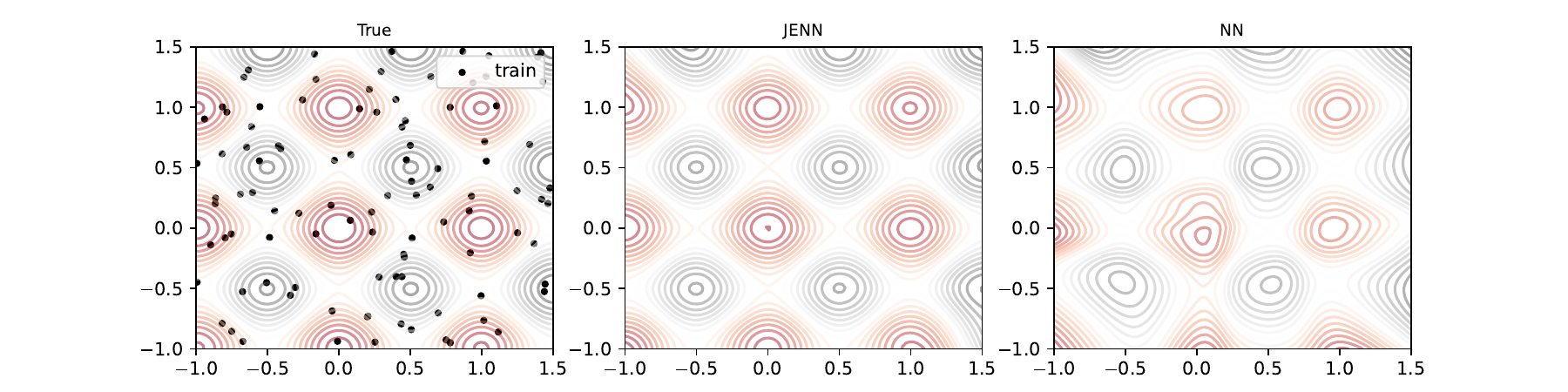}%
    \label{subfig:validation:rastrigin}
}
\caption{Validation results}
\label{fig:validation}
\end{figure*}

Using canonical test functions, Fig.~\ref{fig:validation} compares JENN against a standard Neural Net (NN), taken to be the same neural network with gradient-enhancement turned off (\textit{i.e.} $\gamma=0$). All else being equal, it can be seen that JENN outperforms NN in all cases, reaffirming the general concensus that gradient-enhancement achieves better accuracy with less data. 
In Figs.~\ref{subfig:validation:sinx} and~\ref{subfig:validation:xsinx}, where there are only three and four training samples respectively, NN fails to even capture the correct trend whereas JENN predicts the response almost perfectly. In Fig.~\ref{subfig:validation:rastrigin}, where there are 100 samples, NN correctly captures the trend but contour lines are distorted, implying more samples would be needed to improve prediction. By contrast, under the same conditions, JENN predicts the Rastrigin function~\cite{rastrigin1974systems} almost perfectly. The code to reproduce all results in this section is available online\footnote{code to reproduce available at https://codeocean.com/capsule/9360536/tree}.

\subsection{Run Time}

\begin{figure*}[!t]
    \centering
    \includegraphics[height=1.5in]{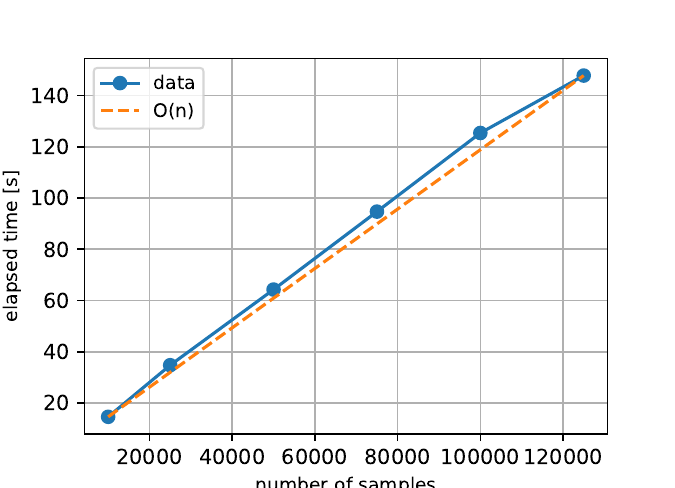}
    \caption{Runtime \textit{vs.} sample size}
    \label{fig:runtime}
\end{figure*}

It was empirically verified that runtime scales linearly as $\mathcal{O}(n)$ with sample size, as shown in Fig.~\ref{fig:runtime}. This plot was obtained using the Rastrigin function \cite{rastrigin1974systems} to generate progressively larger datasets, train a new model, and record runtime on a MacBook Pro (2021).  

\subsection{Surrogate-Based Optimization} 

\begin{figure*}[!t]
    \centering
    \includegraphics[height=3.25in]{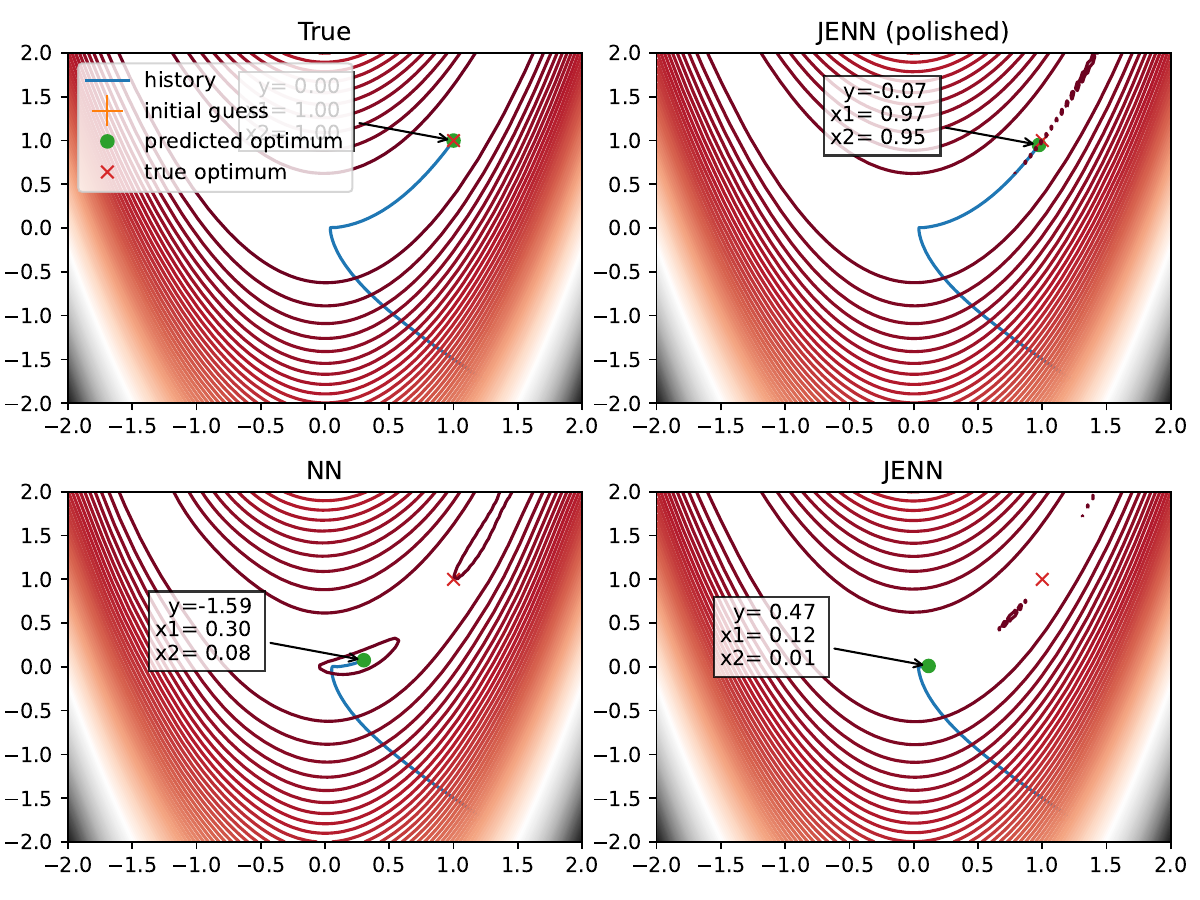}%
    \caption{Optimization history}
    \label{fig:sbo}
\end{figure*}

Surrogate-based optimization is the process of using a fast-running approximation, such as a neural network, to optimize some slow-running function of interest, rather than directly using the function itself. This use case typically arises when the response of interest takes too long to evaluate relative to the time available for analysis. For example, in real-time optimal control, one may wish to consider a neural network to make quick predictions of complicated physics when updating the trajectory~\cite{Sanchez2018}. The purpose of this section is to quantify the benefit of JENN relative to regular neural networks in the context of gradient-based optimization. This will be accomplished by solving the following, constrained optimization problem: 
\begin{align}
    \text{Minimize:} &\quad y = \hat{f}(x_1, x_2) \\
    \text{With Respect To:} &\quad x_1 \in \mathbb{R}, x_2 \in \mathbb{R} \\
    \text{Subject To:} &\quad -2 \le x_1 \le 2 \\ &\quad -2 \le x_2 \le 2
\end{align}
where $\hat{f}$ is a surrogate-model approximation to the Rosenbrock function~\cite{Rosenbrock1960}, which is often used as a benchmark in optimization because the optimum lies in a ``shallow valley'' with near-zero slope that stalls convergence: 
\begin{align}
    f(x_1, x_2) = (1 - x_1)^2 + 100 (x_2 - x_1^2)^2
\end{align}
Results are shown in Fig.\ref{fig:sbo}, which compares optimization convergence histories for the true response and three surrogate-model approximations: NN, JENN, and JENN (polished). Training data is not shown to avoid clutter, but was generated by sampling the true response on a $9\times9$ regular grid plus 100 Latin hypercube points. 


\subsubsection{Results Without Polishing}

Under perfect conditions, one would expect the optimization history to follow the same path for all models, but this is not the case. Neither JENN nor NN reach the true location of this optimum situated at $(1, 1)$ with a value of zero. Carefully studying the contour lines of NN in the region where the Rosenbrock function flattens out, one can observe subtle undulations, likely within the noise of surrogate prediction error, which create artificial local minima that trap the optimizer. This behavior is the Achille's heel of surrogate-based optimization methods. These effects are less pronounced for JENN, but the optimizer gets trapped nonetheless. This finding is not surprising whenever the optimum lies in a \textit{flat} region and approaches zero, since magnitudes are so small they cannot be distinguished from prediction error. 

For a broad set of engineering problems, the results attained by these baseline models may be sufficiently accurate. However, there are situations where optimization accuracy is desirable if not critical, such as aerodynamic shape optimization. In general, surrogate-based optimization tends to struggle any time the magnitude of the slopes in the region of interest lie within prediction noise. This situation is at odds with the needs of gradient-based optimization which requires extreme accuracy near the optimum; accuracy elsewhere is desirable but less critical, provided slopes correctly point in the direction of improvement. An imperative research question therefore arises: how can these competing needs be reconciled? 

\subsubsection{Results With Polishing}

This situation can be ameliorated through ``polishing'' which is uniquely afforded by this framework. Given a trained model, this approach consists of training it further after magnifying regions of interest. This is accomplished by judiciously setting $\gamma$ to prioritize specific training points belonging to that subspace. Specifically, \textit{flat} regions can be magnified by allocating more importance to small slopes during training, using a radial basis functions centered on $\partial y/\partial x = 0$ for example:  
%
\begin{equation}
    \gamma^{(t)} = 1 + \eta \cdot  e^{
    - \left(\epsilon \frac{\partial y_k}{\partial x_j}^{(t)}\right)^2
    } ~\forall ~ t, k, j
\end{equation}
where $\eta$ and $\epsilon$ are hyperparameters that control the radius of influence of a point. They were set to $\eta=1000$ and $\epsilon=0.1$ in the results shown. Far away from any point $x^{(t)}$ where $\partial y / \partial x^{(t)}=0$, the weights are $\gamma^{(t)}\approx1$ whereas near those points $\gamma^{(t)}\approx1001$. This has the desired effect of disproportionately emphasizing regions of small slope ${\partial y_k}/{\partial x_j}$ during training up to a factor of 1000 at the expense of deprioritizing other regions, which is why it is necessary to start from an already trained model. Results are shown in Fig.~\ref{fig:sbo} for the Rosenbrock test case, where it can be seen that JENN (polished) now recovers the true optimum almost exactly.    

\subsection{Noisy Partials}

We now consider the effect of noisy partials. Many practical design applications rely on legacy software suites during early design phases, such as missile DATCOM~\cite{DATCOM}, for which analytical or automatically differentiated partials are not available. In this case, the only recourse for computing derivatives is finite difference. However, despite their attractive simplicity, it has been shown that finite difference methods are not as accurate as their counterparts~\cite{MartinsHwang2013}. It is therefore natural to wonder: how accurate must partials be in order to take advantage of JENN in design applications where finite difference might be unavoidable? 
\begin{figure}[h!]
  \centering
  \includegraphics[width=2.0in]{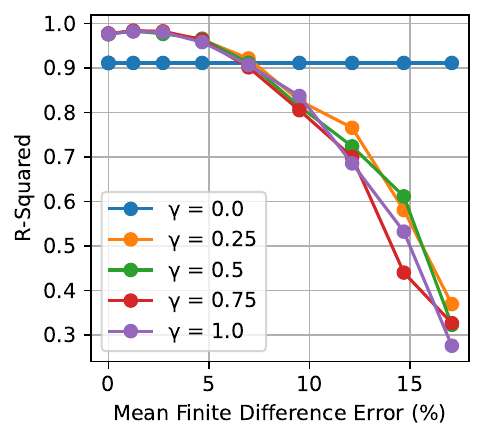}
  \caption{Predictive quality as a function of noisy partials}
  \captionsetup{justification=centering}
  \label{fig:noisy_partials}
\end{figure}

The answer to this question is shown by example in Fig.~\ref{fig:noisy_partials} using the Rastrigin function as a testbed once again, where $\gamma=0$ corresponds to baseline results for a standard neural network (gradient-enhancement turned off). Partial derivatives of the training data were obtained using finite difference with progressively increasing step sizes. The corresponding mean finite difference error shown was computed against true analytical partials. It can clearly be seen that beyond seven percent error, the benefit of gradient-enhancement vanishes. In fact, beyond that threshold, partials are too inaccurate to provide useful information: accounting for erroneous partials is worse than not accounting for partials at all. These results are welcome, as they demonstrate that the usefulness of gradient-enhanced methods is not restricted to perfectly accurate partials; even noisy partials can be useful (within a reasonable threshold).  

These results further show that there is no benefit to setting $\gamma$ to values other than zero or one; it does not mitigate the effect of noisy partials.  For most applications, the hyperparameter $\gamma$ should therefore be treated as a binary scalar variable to turn gradient-enhancement on ($\gamma=1$) or off ($\gamma=0$); except in special situations such as incomplete partials or polishing, as previously illustrated, which call for a distribution of $\gamma$ in that case. Incomplete partials refers to the scenario where partials are only available for some inputs, in which case $\gamma$ can be used to turn gradient-enhancement off for the missing partials, \textit{i.e.} resulting in a tensor of zeros and ones as illustrated next in airfoil application section.    
\section{Application: Airfoil Shape Optimization}

This section quantifies the benefit of JENN for aerodynamic shape optimization through applied example. This is accomplished by optimizing the shape of a NACA 0012 airfoil in inviscid, transonic flow and comparing results to \textit{status quo} neural networks. 

\subsubsection{Problem Description}

To ensure reproducibility, the open-source software SU2~\cite{Economon2016} was selected for computational fluid dynamics. The design problem is taken directly from their website\footnote{\url{https://su2code.github.io/tutorials/Inviscid_2D_Unconstrained_NACA0012/}}, modified to hold lift coefficient fixed (instead of angle of attack), and airfoil geometry is parameterized using fourteen Hicks-Henne ``bump'' functions~\cite{Hicks1978}. The goal is to minimize drag $C_d$ at some desired flight condition, specified by Mach number $M$ and lift coefficient $C_l$: 
\begin{align}
    \text{Minimize:} &\quad C_d = f(x_0, \dots, x_{13}, M, C_l) \label{eq:fobj}\\
    \text{With Respect To:} &\quad x_0, \dots, x_{13} \label{eq:vars}\\
    \text{Subject To:} 
    &\quad -0.01 \le x_i \le 0.01 ~\forall ~ i \in \{0, \dots, 13\}  \label{eq:cons}
\end{align}
where $x_0, \dots, x_{13}$ denote the Hicks-Henne shape parameters. The flight condition parameters $(M, C_l)$ determine the function $f$, which effectively defines a different instance of the optimization problem. Each instance took several minutes to solve using adjoint methods on a modern laptop. 

\subsubsection{Statistical Regression}

\begin{figure*}[!t]
    \centering
    \includegraphics[width=\linewidth]{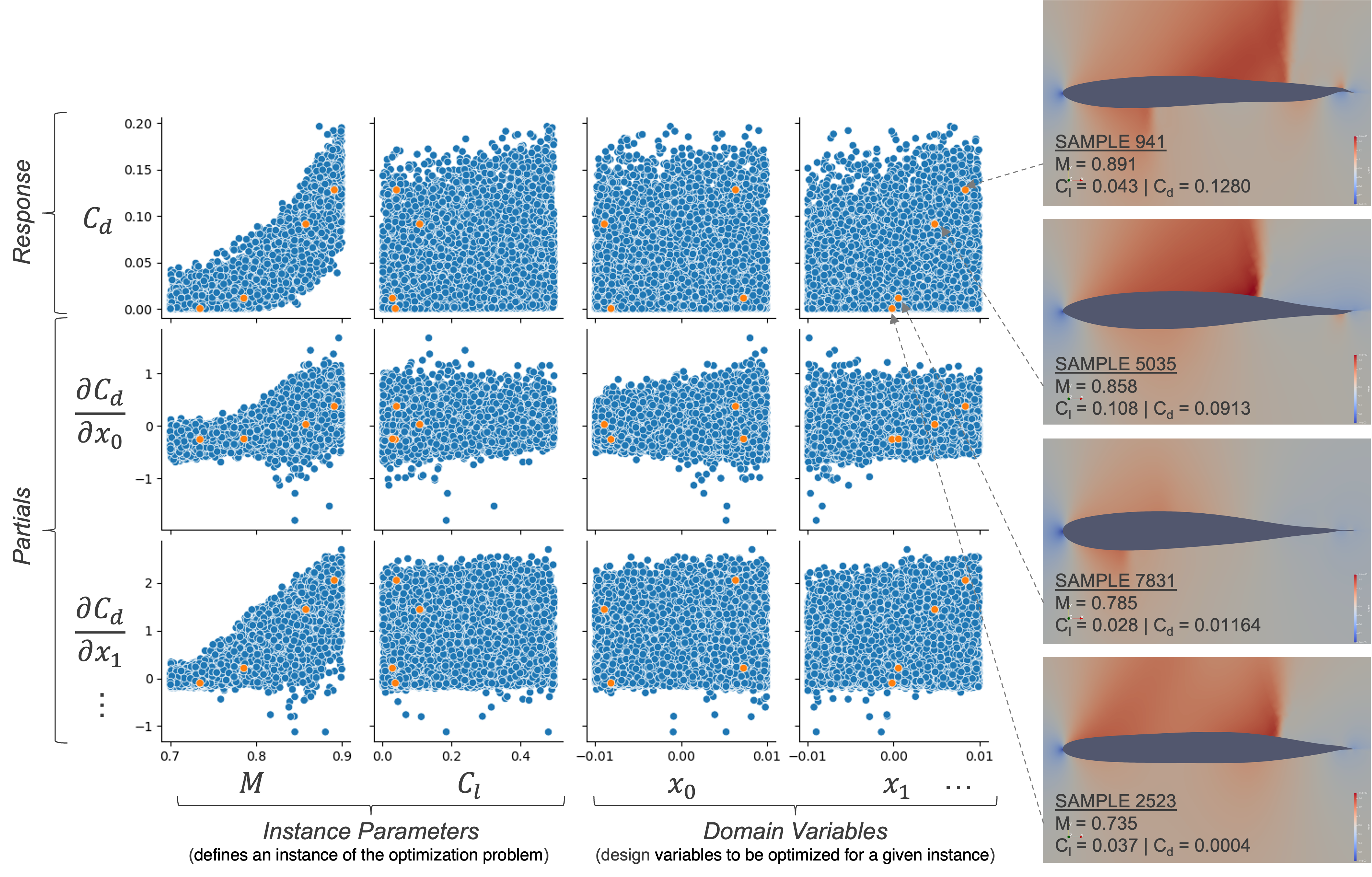}
    \caption{Sample data using Latin hypercube sampling}
    \label{fig:sample-data}
\end{figure*}

Our goal is to accelerate numerical optimization by developing an optimal proxy, which can be used to near instantaneously shape an airfoil at any flight condition. In the proposed approach, this is accomplished by replacing the objective function $f$ by an equivalent approximation $\hat{f}$, taken to be a neural network. To do so, one must therefore sample a 16-dimensional input space composed of 14 shape variables $x_0, \dots, x_{13}$ and two flight conditions $0.7 \le M \le 0.9$ and $0 \le C_l \le 0.5$ to generate training data. Sample results are shown in Fig.~\ref{fig:sample-data} for a subset of the design space. A total of 10,000 Latin hypercube samples were evaluated.  

\begin{figure*}[!t]
\centering
\subfloat[R-squared]{
    \includegraphics[width=0.48\textwidth]{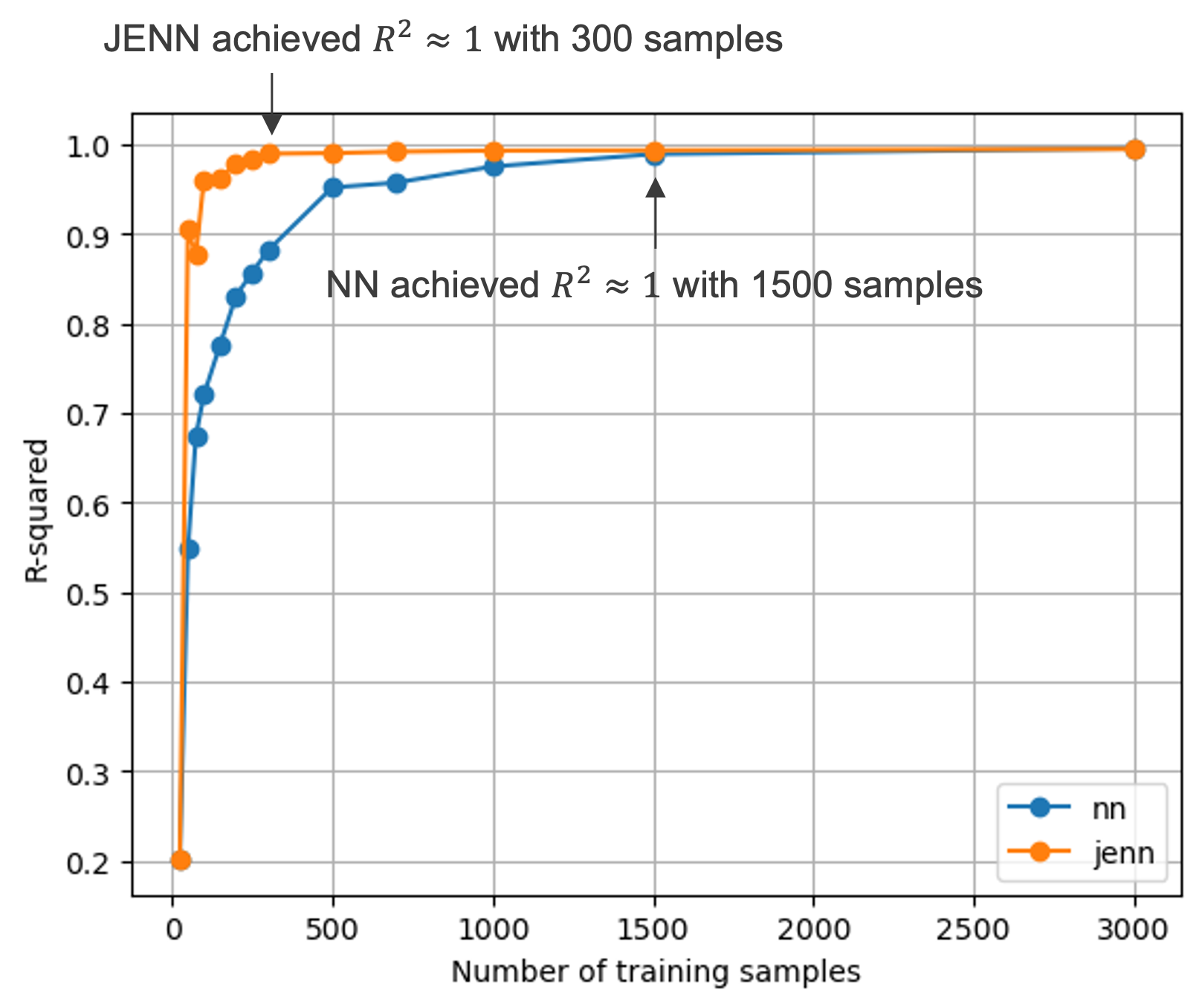}%
    \label{subfig:trades:rsquared}
}
\subfloat[Error standard deviation]{
    \includegraphics[width=0.48\textwidth]{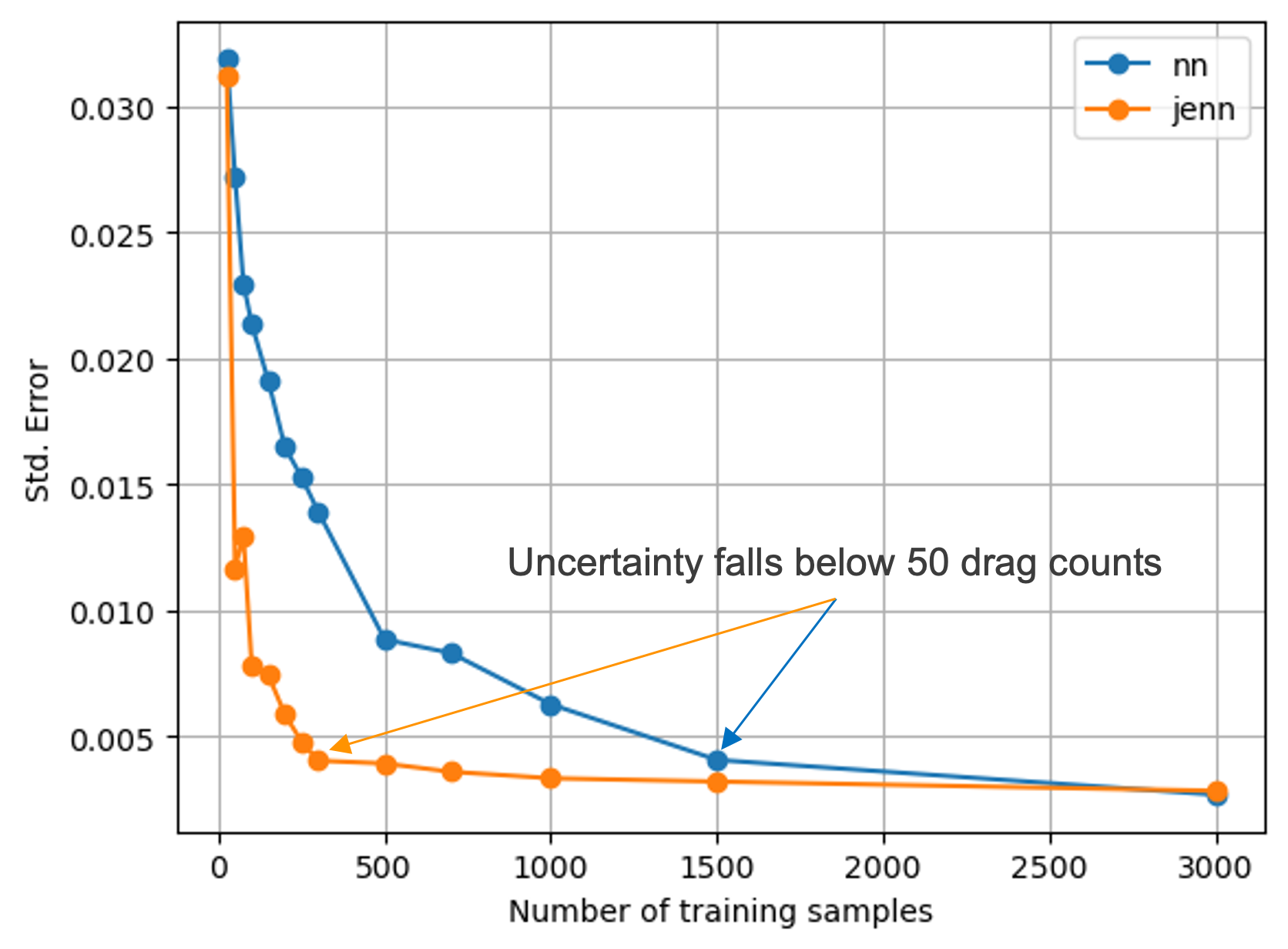}%
    \label{subfig:trades:stdev}
}
\caption{$C_d$ prediction accuracy as a function of training samples}
\label{fig:trades}
\end{figure*}

Upon collecting samples, the data was used to determine how many training points were needed to achieve good prediction accuracy using JENN and regular Neural Networks (NN). This was accomplished by holding back a subset of the data for training and using the rest for testing. The neural net architecture (5 hidden layers of 16 neurons each) and hyperparameter settings were identical for both neural networks, except for gradient enhancement which was turned on or off accordingly. The results are shown in Fig.~\ref{fig:trades} using test data to plot the R-squared value as function of the number of training samples. Results unambiguously show that JENN achieves the same prediction accuracy of 99\% using five times fewer samples, re-affirming the conclusion by other authors that gradient-enhanced methods reduce the amount of data required for good prediction, even on practical engineering problems. 

Using the results of Fig.~\ref{fig:trades}, it is possible to identify the minimum number of points needed to train an accurate neural net under both paradigms for this problem, as indicated by the arrows. Associated convergence properties and goodness of fit were verified to be satisfactory but not reported for conciseness, except for Table~\ref{tab:partials_rsquare} which compares the R-squared value of the partials in order to emphasize that JENN predicts them more accurately.

\begin{table}[h]
    \centering
    \resizebox{\columnwidth}{!}{
    \begin{tabular}{lrrrrrrrrrrrrrr}
        \hline
          \textbf{$R^2 $} & \textbf{$\partial x_0$} & \textbf{$\partial x_1$} & \textbf{$\partial x_2$} & \textbf{$\partial x_3$} & \textbf{$\partial x_4$} & \textbf{$\partial x_5$} & \textbf{$\partial x_6$} & \textbf{$\partial x_7$} & \textbf{$\partial x_8$} & \textbf{$\partial x_9$} & \textbf{$\partial x_{10}$} & \textbf{$\partial x_{11}$} & \textbf{$\partial x_{12}$} & \textbf{$\partial x_{13}$} \\
        \hline
        NN & -0.69 & 0.79 & 0.79 & 0.68 & 0.32 & 0.26 & 0.32 & -0.95 & 0.73 & 0.80 & 0.77 & 0.59 & 0.49 & -0.10 \\
        JENN & 0.37  & 0.90 & 0.89 & 0.78 & 0.59 & 0.18 & 0.34 & 0.29  & 0.85 & 0.88 & 0.87 & 0.77 & 0.59 & 0.13  \\
        \hline
    \end{tabular}
    }
    \caption{Comparison of the R-squared value of the partials across models}
    \label{tab:partials_rsquare}
\end{table}

\begin{figure*}[!t]
\centering
\subfloat[$C_d$ vs. $M$]{
    \includegraphics[width=0.23\textwidth]{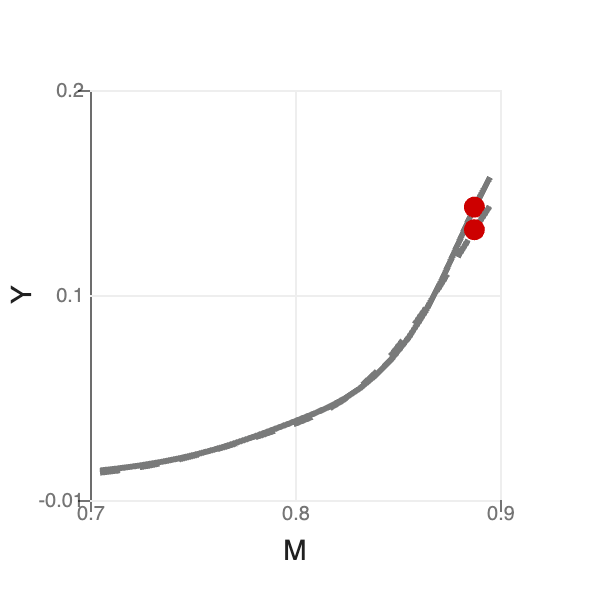}%
    \label{subfig:profiler:CD_vs_M}
}
\subfloat[$C_d$ vs. $C_l$]{
    \includegraphics[width=0.23\textwidth]{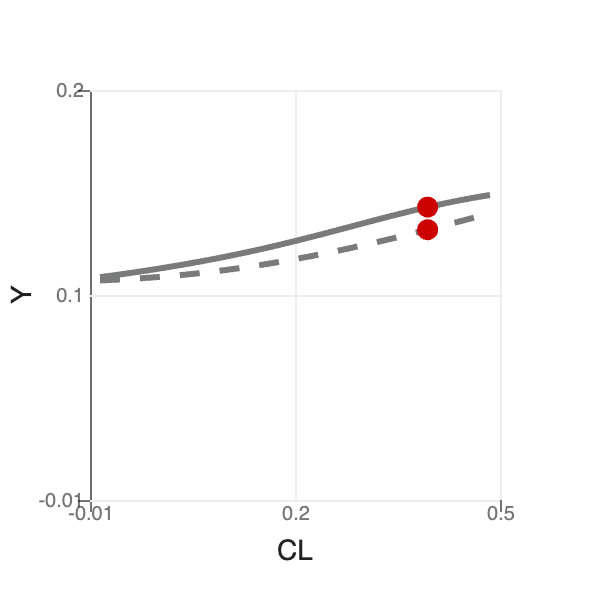}%
    \label{subfig:profiler:CD_vs_CL}
}
\subfloat[$C_d$ vs. $x_0$]{
    \includegraphics[width=0.23\textwidth]{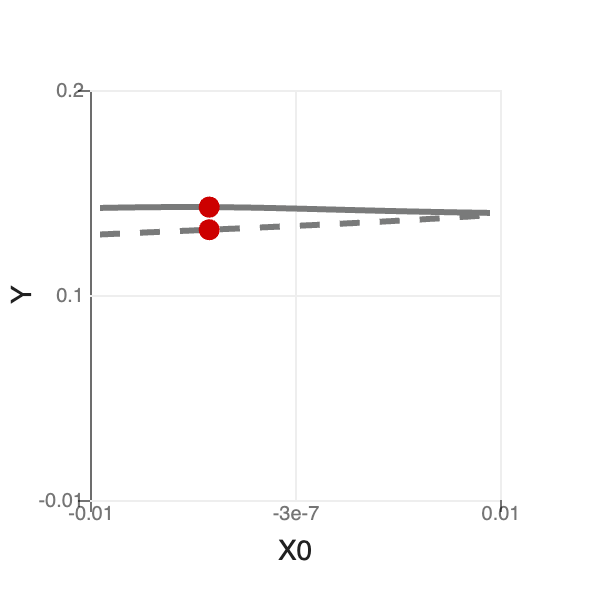}%
    \label{subfig:profiler:CD_vs_X0}
}
\subfloat[$C_d$ vs. $x_1$]{
    \includegraphics[width=0.23\textwidth]{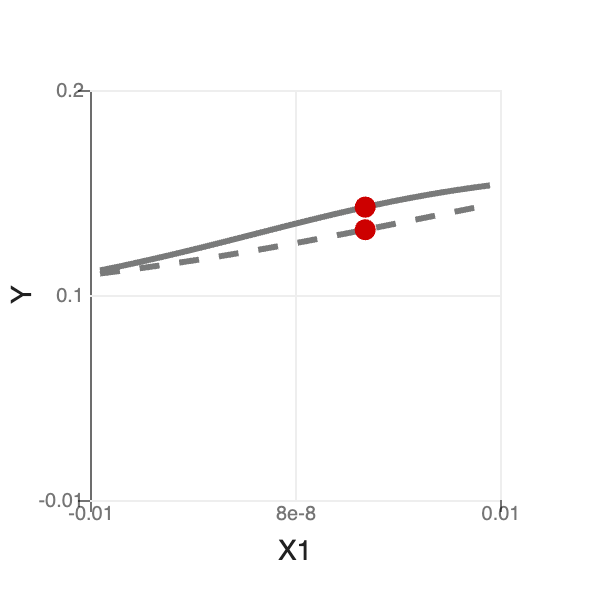}%
    \label{subfig:profiler:CD_vs_X1}
}
\\
\subfloat[$C_d$ vs. $x_2$]{
    \includegraphics[width=0.23\textwidth]{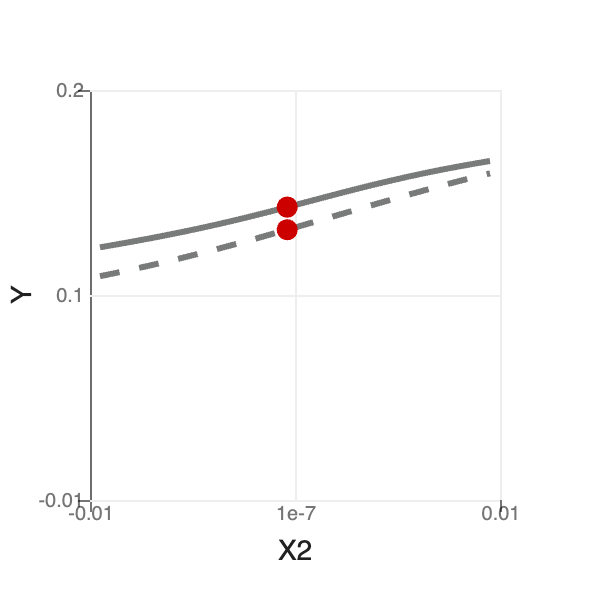}%
    \label{subfig:profiler:CD_vs_X2}
}
\subfloat[$C_d$ vs. $x_3$]{
    \includegraphics[width=0.23\textwidth]{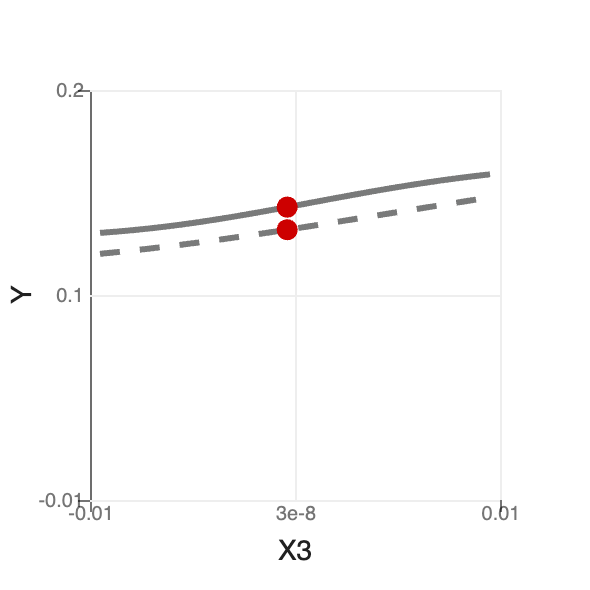}%
    \label{subfig:profiler:CD_vs_X3}
}
\subfloat[$C_d$ vs. $x_4$]{
    \includegraphics[width=0.23\textwidth]{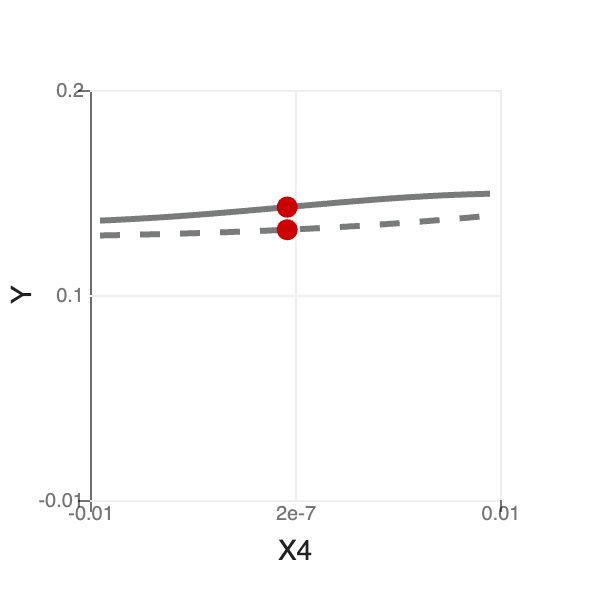}%
    \label{subfig:profiler:CD_vs_X4}
}
\subfloat[$C_d$ vs. $x_5$]{
    \includegraphics[width=0.23\textwidth]{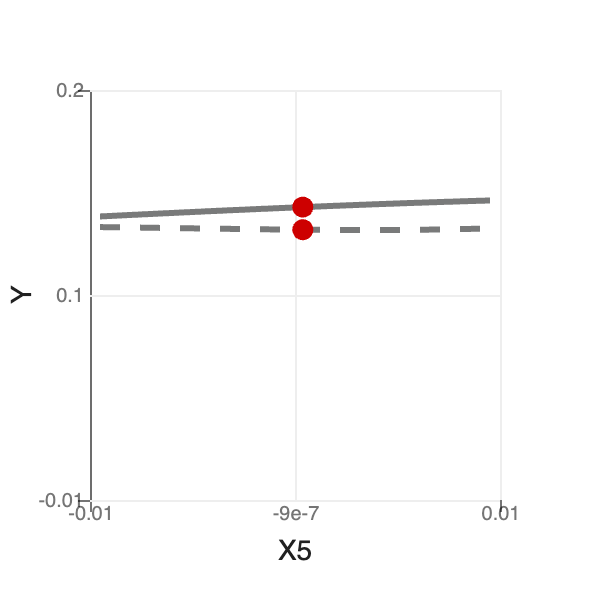}%
    \label{subfig:profiler:CD_vs_X5}
}
\\
\subfloat[$C_d$ vs. $x_6$]{
    \includegraphics[width=0.23\textwidth]{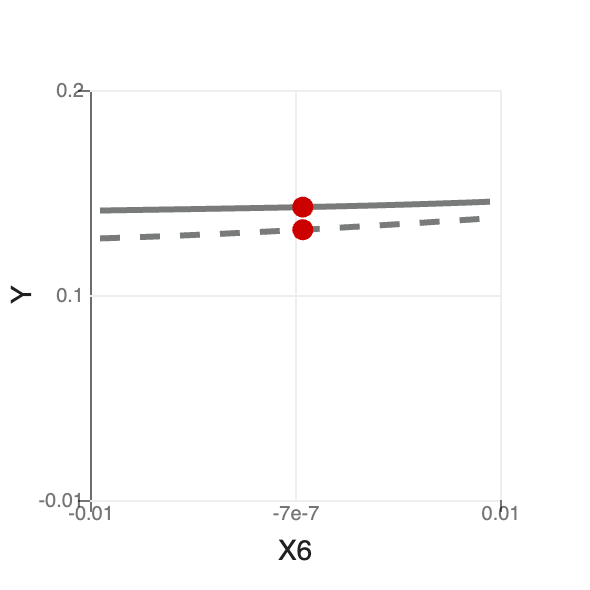}%
    \label{subfig:profiler:CD_vs_X6}
}
\subfloat[$C_d$ vs. $x_7$]{
    \includegraphics[width=0.23\textwidth]{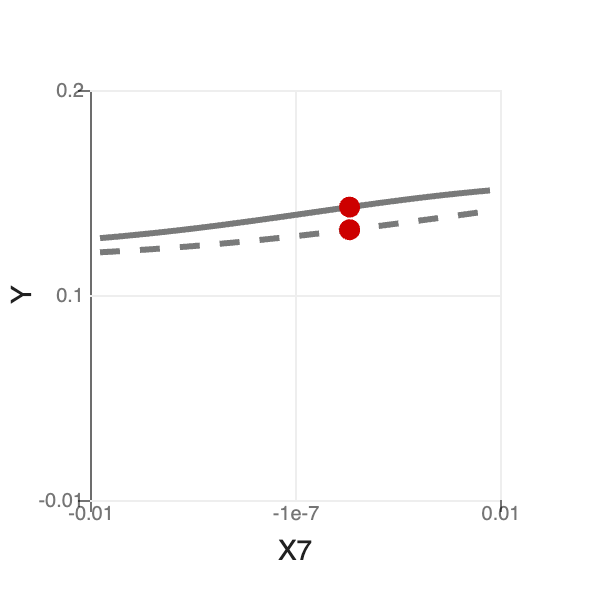}%
    \label{subfig:profiler:CD_vs_X7}
}
\subfloat[$C_d$ vs. $x_8$]{
    \includegraphics[width=0.23\textwidth]{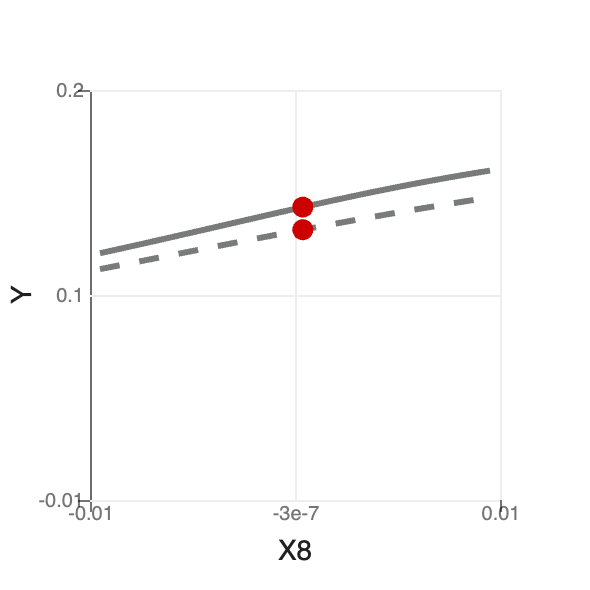}%
    \label{subfig:profiler:CD_vs_X8}
}
\subfloat[$C_d$ vs. $x_9$]{
    \includegraphics[width=0.23\textwidth]{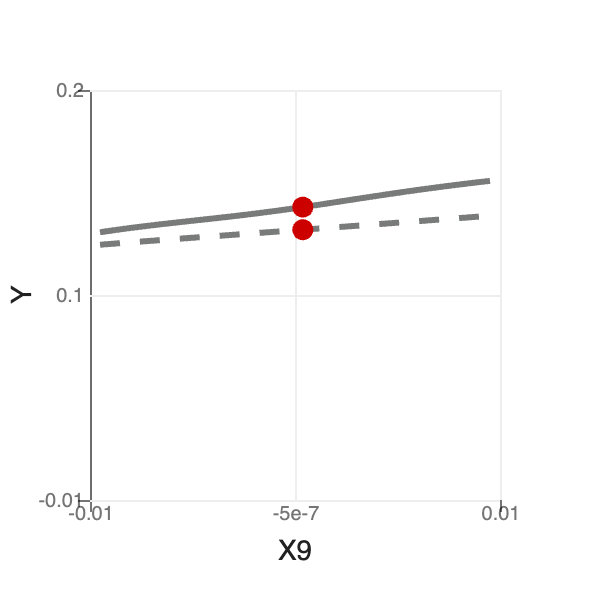}%
    \label{subfig:profiler:CD_vs_X9}
}
\\
\subfloat[$C_d$ vs. $x_{10}$]{
    \includegraphics[width=0.23\textwidth]{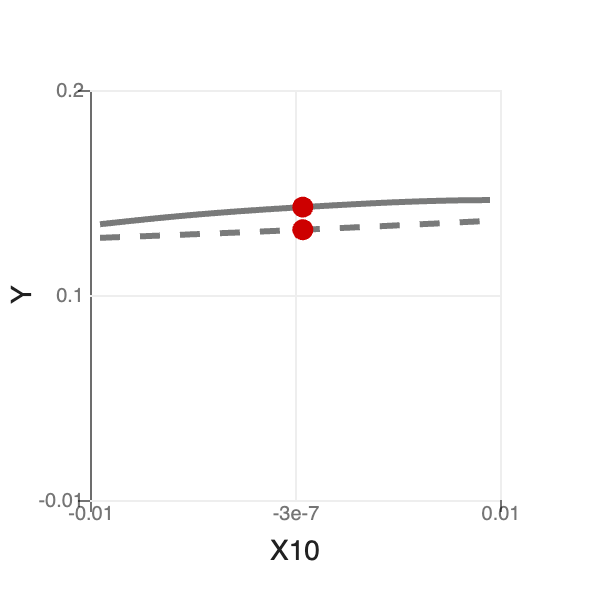}%
    \label{subfig:profiler:CD_vs_X10}
}
\subfloat[$C_d$ vs. $x_{11}$]{
    \includegraphics[width=0.23\textwidth]{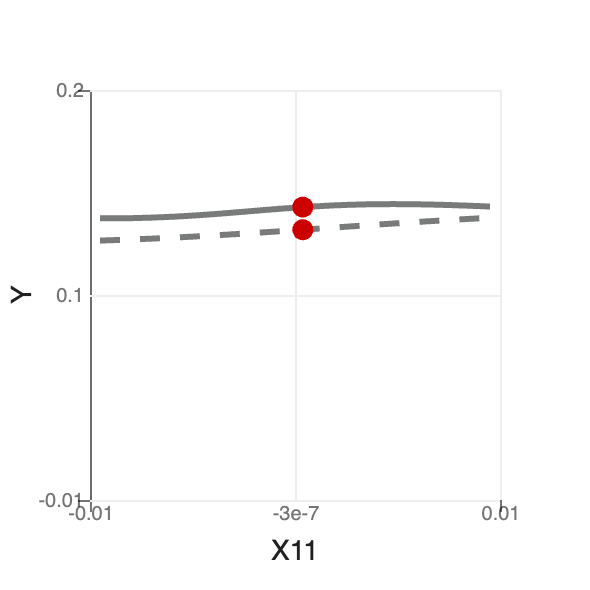}%
    \label{subfig:profiler:CD_vs_X11}
}
\subfloat[$C_d$ vs. $x_{12}$]{
    \includegraphics[width=0.23\textwidth]{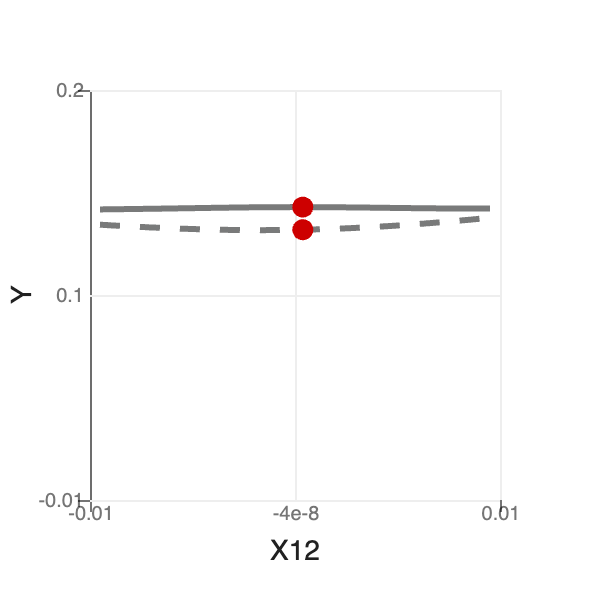}%
    \label{subfig:profiler:CD_vs_X12}
}
\subfloat[$C_d$ vs. $x_{13}$]{
    \includegraphics[width=0.23\textwidth]{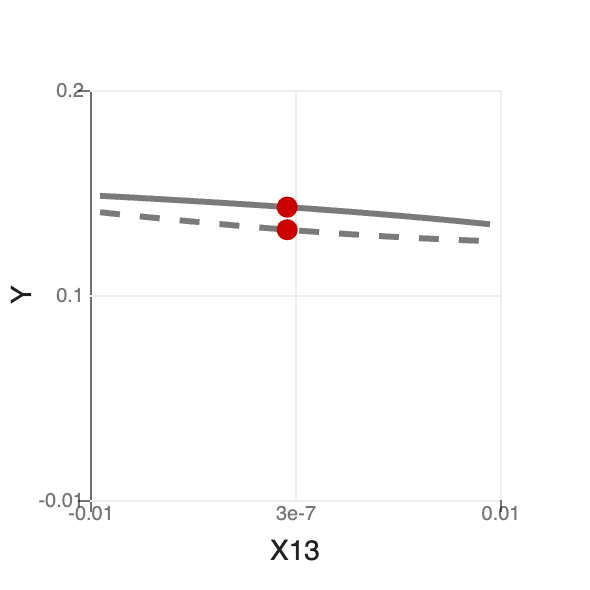}%
    \label{subfig:profiler:CD_vs_X13}
}
\caption{Sensitivity profiles for \textit{status quo} NN (solid line) and JENN (dashed line)}
\label{fig:profiler}
\end{figure*}

Corresponding sensitivity profiles are shown in Fig.~\ref{fig:profiler} for all variables, in order to show the topological differences that resulted from gradient enhancement. Overall, it was found that JENN and NN exhibit excellent agreement in the subsonic regime, but that discrepancies start to appear in transonic flow at higher Mach numbers, where shock waves are present, presumably creating stronger sensitivity to shape variables that control the region where shocks tend to appear. As a result, the surrogate model learns to predict sensitivities more accurately in some regions than others, which can be seen from the prediction profiles. Although there is good agreement in general, notice that NN slopes occasionally point in a different direction in regions that tend to be flatter. This could have subtle consequences during optimization.

Finally, it should be noted that SU2 did not provide partial derivatives for $\partial C_d / \partial M$ or $\partial C_d / \partial C_l$, which implies incomplete gradient information. This is particularly noteworthy because most engineering design applications will fall in this scenario: partials are only available for some variables. Yet, by virtue of the framework afford by JENN, such missing information is easily overcome by simply setting $\gamma=0$ for the missing partials. This represents a simple but important practical improvement over similar reported work. 

\subsubsection{Surrogate-Based Optimization}

The surrogate models $\hat{f}$ developed in the previous subsection were substituted for $f$ in the optimization problem described by Eqs.~\ref{eq:fobj}--\ref{eq:cons}, which was solved using the SLSQP algorithm from Scipy~\cite{2020SciPy-NMeth}. Fig.~\ref{fig:sbo} compares  final solutions against the initial geometry using different approaches. All solutions successfully recovered a supercritical airfoil shape, as expected, but adjoint-based optimization took several minutes, whereas surrogate-based optimization took less than a second, setting the cost of generating training data aside. However, there are some observable differences. In particular, notice the presence of a weak but unwanted secondary shock wave in the answer obtained using a regular NN. This is necessarily attributed to small differences in predicted partial derivatives, since that is the only difference between JENN and a regular NN. This underlines the foreseeable conclusion that gradient-enhanced methods ameliorate surrogate-based optimization somewhat whenever gradient-descent algorithms are used. Overall, predictions remained comparable, but the cost of generating data was significantly less for JENN.     

\begin{figure*}[!t]
\centering
\subfloat[Initial Geometry: $C_d=0.0346$]{
    \includegraphics[width=0.48\textwidth]{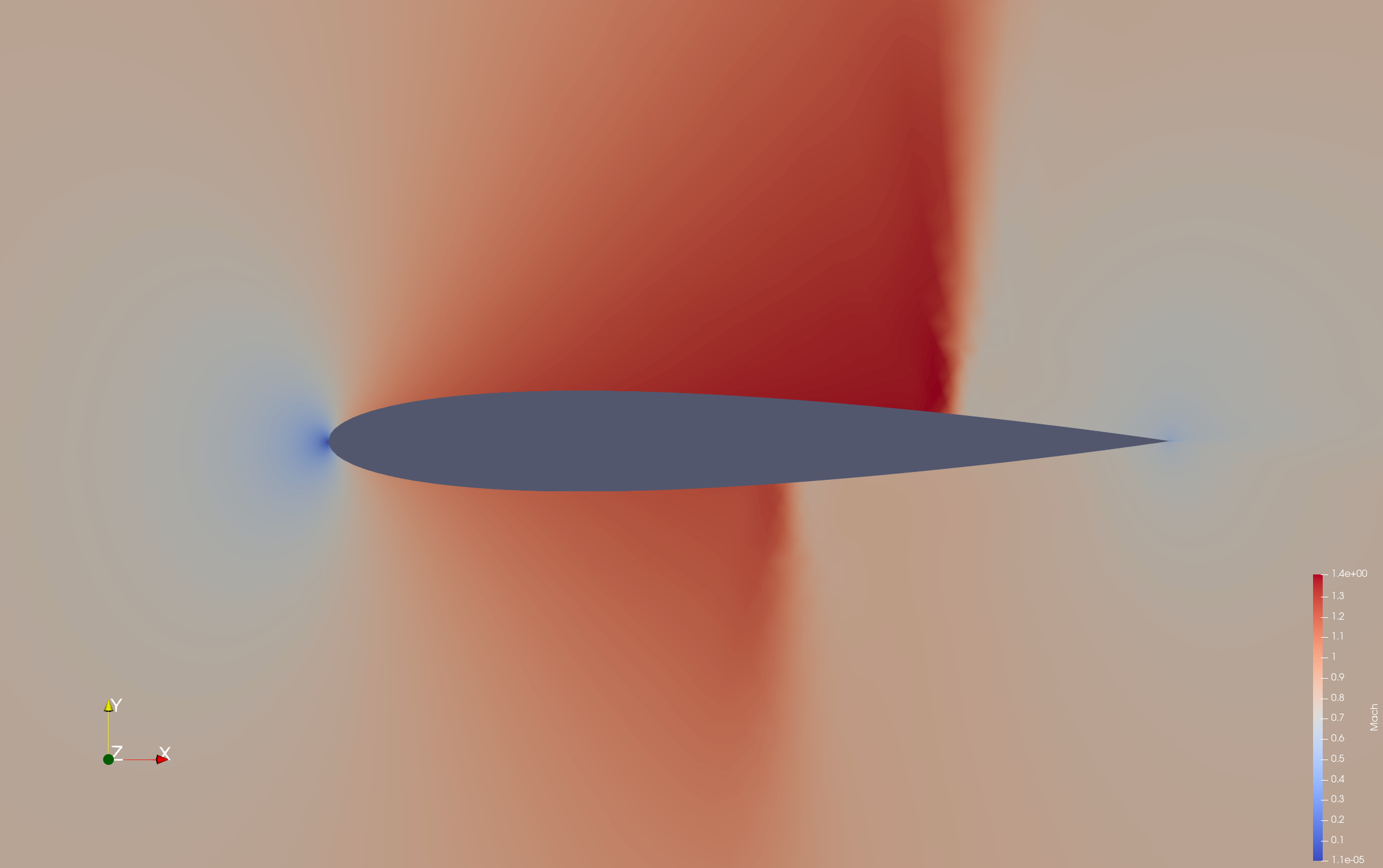}%
    \label{subfig:sbo:guess}
}
\subfloat[Adjoint: $C_d=0.0045$]{
    \includegraphics[width=0.48\textwidth]{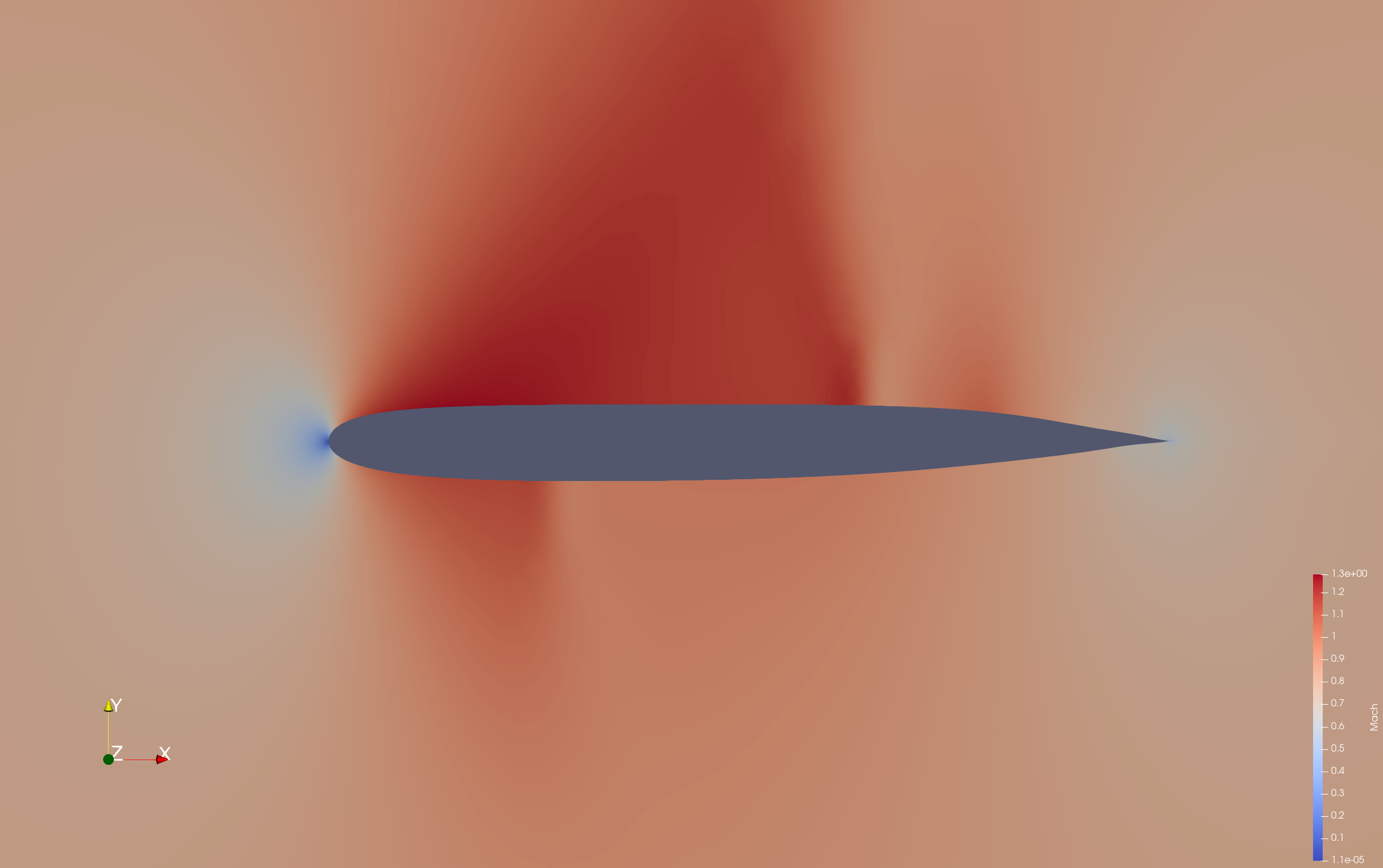}%
    \label{subfig:sbo:adjoint}
}
\\
\subfloat[NN (1500 training data): $C_d=0.0045$]{
    \includegraphics[width=0.48\textwidth]{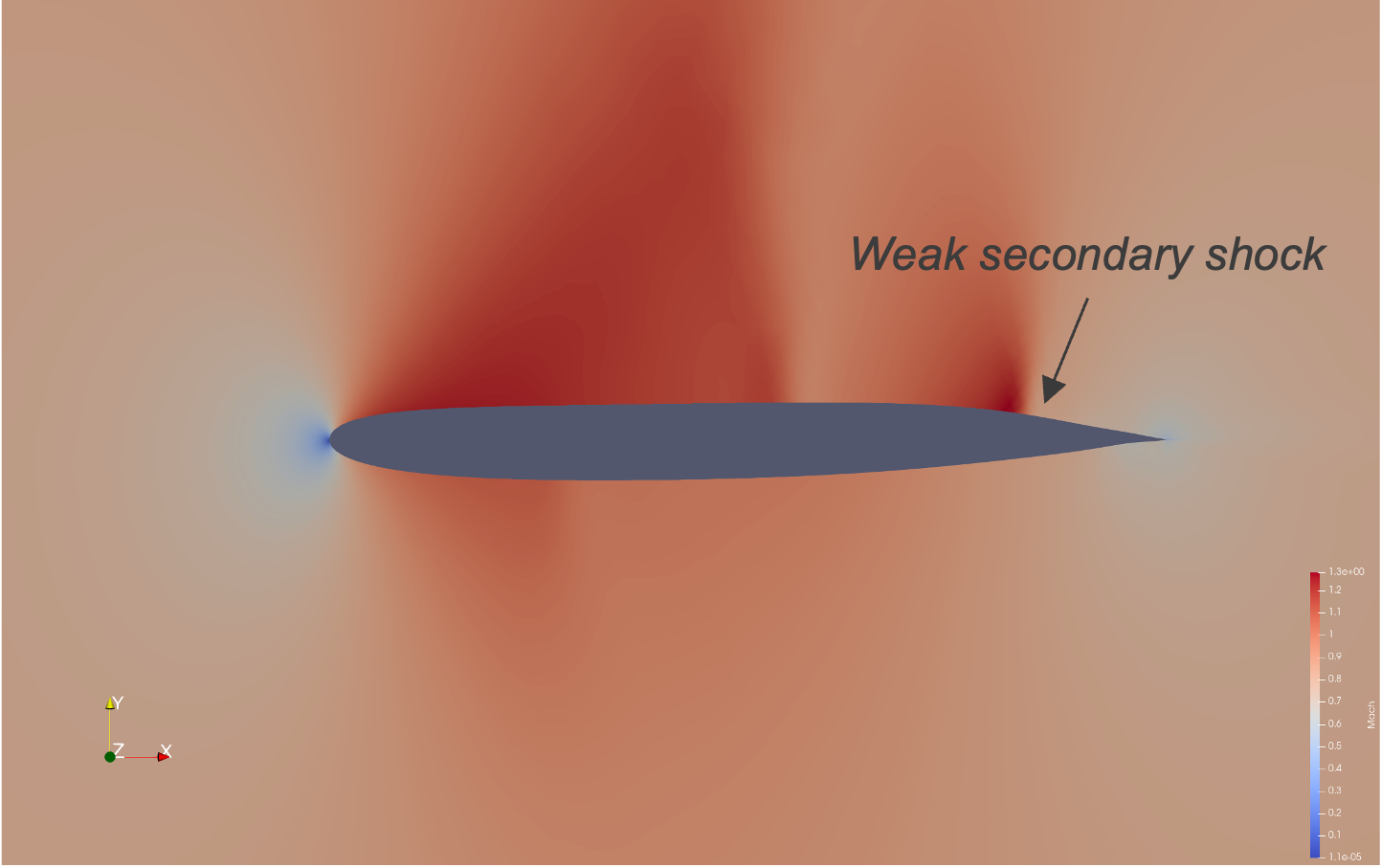}%
    \label{subfig:sbo:ann}
}
\subfloat[JENN (300 training data): $C_d=0.0023$]{
    \includegraphics[width=0.48\textwidth]{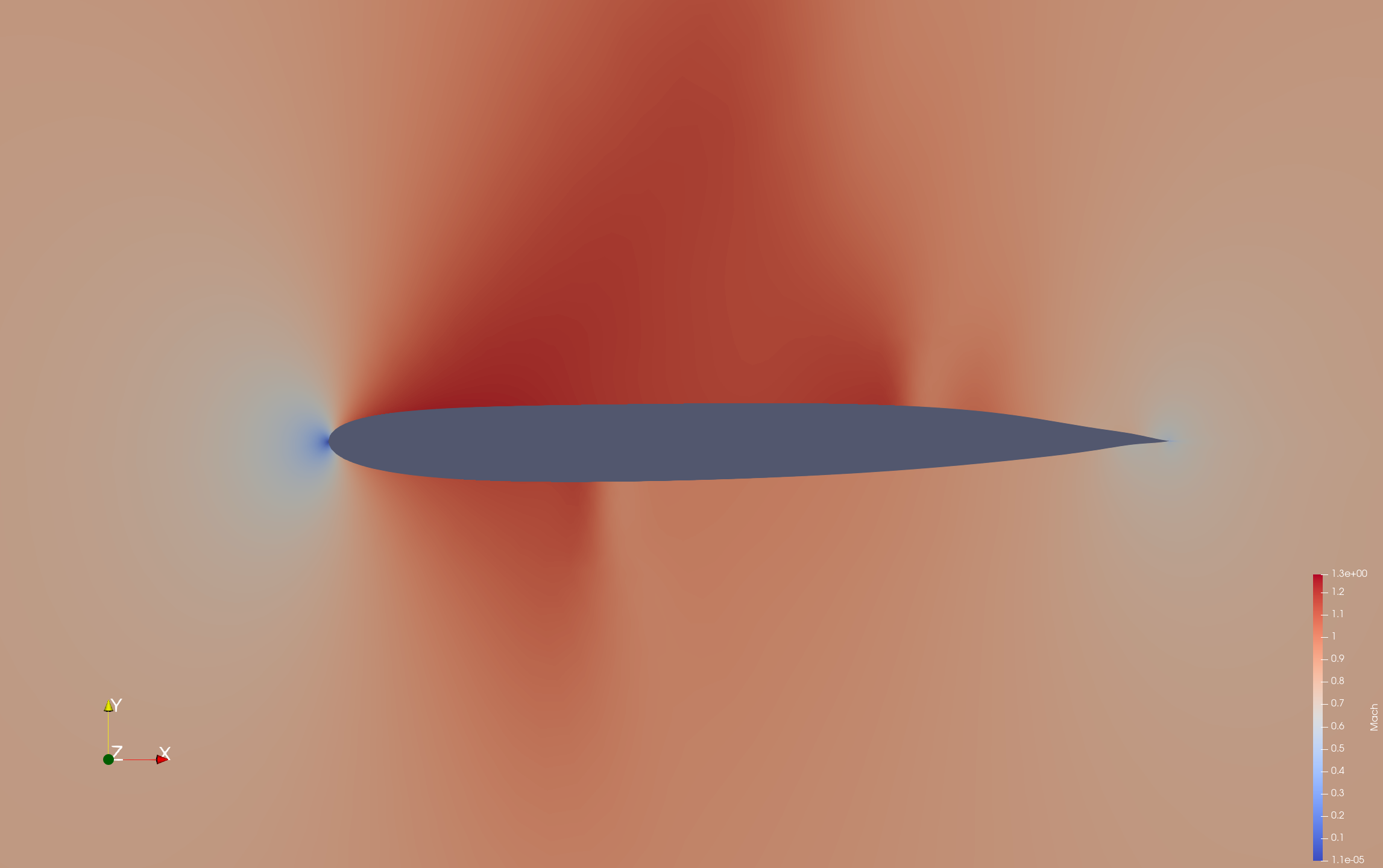}%
    \label{subfig:sbo:gnn}
}
\caption{Inviscid NACA 0012 shape optimization results at $M=0.83$ and $C_l=0.28$}
\end{figure*}
\section{Conclusion}

In summary, this research developed JENN, which are densely connected multi-layer perceptrons that predict gradients more accurately, successfully validated the implementation against canonical test functions, and applied it to a practical airfoil shape optimization problem. All results support the general consensus that gradient-enhanced methods require less training data than their \textit{status quo} neural network counterparts and achieve better accuracy, provided the cost of obtaining training data partials is not prohibitive in the first place. 

It was further shown that JENN outperforms standard neural networks in the context of surrogate-based optimization, but that surrogate-based methods are more prone to getting trapped in artificial local minima, whenever the true optimum lies in flat regions and approaches zero. This was apparent for the Rosenbrock test function, but also for the airfoil design problem where the regular NN solution exhibited a weak secondary shockwave that was not present with JENN. Overall, JENN tends to achieve optimal solutions that are in better agreement with the true optimum. 

Furthermore, through a process dubbed \textit{polishing}, uniquely afforded by this framework, JENN enables salient regions of interest to be magnified through judicious use of hyperparameters. This feature provides an additional \textit{knob} for improving surrogate-based optimization accuracy in situations that call for it, as demonstrated with the Rosenbrock example. The same mechanism also enables missing partials to be ignored during training by setting $\gamma=0$ for those partials, such that gradient-enhancement can be leveraged even in situations when information is incomplete. This is likely to be a common scenario for many engineering design problems, as was the case for the airfoil design example where $\partial C_d / \partial M$ and $\partial C_d / \partial C_l$ were missing. This simple mechanism is likely to be of important practical value. 

In closing, despite its advantages over \textit{status quo} neural networks, JENN does not overcome the curse of dimensionality; although it alleviates it by reducing the number of samples needed. Sampling remains necessary to generate training data, and so the practical utility of gradient-enhanced methods is likely to be restricted to problems where intrinsic dimensionality is either small to start with or can be recovered using elaborate dimensionality reduction techniques~\cite{BerguinMavris2015, BerguinRancourtMavris2015, Constantine2015}. To be adopted at industrial scales, new techniques are needed to avoid high-dimensional sampling whilst retaining the benefit of optimal proxies for accelerating numerical optimization. 



\section*{Acknowledgments}

The supporting code for this work used the exercises by Professor Andrew~\cite{Ng2017} as a starting point, but fundamentally change the formulation to handle Jacobian enhancement. The primary author, who was then a student, would therefore like to thank Professor Ng for offering the fundamentals of deep learning on Coursera, which masterfully explained what can be considered a challenging and complicated subject in an easy way to understand.


\newpage

\appendix
\section{Key Equations in Vectorized Form}
\label{app:vectorized-equations}

This section provides vectorized equations to facilitate re-implementation or as a companion guide to the JENN library. 

\subsection{Data Structures}

Using capital letters, notation can be extended to matrix notation to account for all $m$ training examples, starting with training data inputs and outputs:  
\begin{align}
    \boldsymbol{X} 
   &=
   \left[
   \begin{matrix}
   x_1^{(1)} & \dots & x_1^{(m)} \\
   \vdots & \ddots & \vdots \\
   x_{n_x}^{(1)} & \dots & x_{n_x}^{(m)} \\
   \end{matrix}
   \right]
   \in 
   \mathbb{R}^{n_x \times m}
   \\ 
   \boldsymbol{Y} 
   &=
   \left[
   \begin{matrix}
   y_1^{(1)} & \dots & y_1^{(m)} \\
   \vdots & \ddots & \vdots \\
   y_{n_y}^{(1)} & \dots & y_{n_y}^{(m)} \\
   \end{matrix}
   \right]
   \in 
   \mathbb{R}^{n_y \times m}
\end{align}
The associated Jacobians are formatted as: 
\begin{align}
    \boldsymbol{J} 
   =
   \left[
   \begin{matrix}
   {\left[
   \begin{matrix}
   \dfrac{\partial y_1}{\partial x_1}^{(1)} & \dots & \dfrac{\partial y_1}{\partial x_{1}}^{(m)}  \\
   \vdots & \ddots & \vdots \\
   \dfrac{\partial y_{1}}{\partial x_{n_x}}^{(1)} & \dots & \dfrac{\partial y_{1}}{\partial x_{n_x}}^{(m)}  \\
   \end{matrix}
   \right]}
   \\ 
   \vdots 
   \\ 
   {\left[
   \begin{matrix}
   \dfrac{\partial y_{n_y}}{\partial x_1}^{(1)} & \dots & \dfrac{\partial y_{n_y}}{\partial x_{1}}^{(m)}  \\
   \vdots & \ddots & \vdots \\
   \dfrac{\partial y_{n_y}}{\partial x_{n_x}}^{(1)} & \dots & \dfrac{\partial y_{n_y}}{\partial x_{n_x}}^{(m)}  \\
   \end{matrix}
   \right]}
   \end{matrix}
   \right]
   \in
   \mathbb{R}^{n_y \times n_x \times m}
\end{align}

Note that the hyperparameters $\beta, \gamma$ can either be scalars or real arrays of the same shape as ${\bf Y}$ and ${\bf J}$, respectively. The shape of the neural network parameters to be learned (\textit{i.e.} weights and biases) is independent of the number of training examples. They depend only on layer sizes and are denoted as: 
\begin{align}
{\bf W}^{[l]} &= \left[ \begin{matrix} 
w_{11}^{[l]} & \dots & w_{1n^{[l-1]}}^{[l]} \\ 
\vdots & \ddots & \vdots \\ 
w_{n^{[l]}1}^{[l]} & \dots & w_{n^{[l]}n^{[l-1]}}^{[l]} 
\end{matrix} \right]  
~ \in \mathbb{R}^{n^{[l]} \times n^{[l-1]}}
\\ 
\boldsymbol{b}^{[l]} &= \left[ \begin{matrix} 
b_1^{[l]} \\ 
\vdots \\ 
b_{n^{[l]}}^{[l]} 
\end{matrix} \right]
~ \in \mathbb{R}^{n^{[l]} \times 1}
\quad \forall \quad 1 < l \le L
\end{align}
However, hidden layer activations do depend on the number of examples and are stored as: 
\begin{alignat}{3}
{\bf A}^{[l]} &= \left[ \begin{matrix} 
a_1^{(1)} & \dots & a_1^{(m)~} \\ 
\vdots & \ddots & \vdots \\ 
a_{n^{[l]}}^{(1)} & \dots & a_{n^{[l]}}^{(m)~} 
\end{matrix} \right]  
~ \in \mathbb{R}^{n^{[l]} \times m}
\\
{\bf Z}^{[l]} &= \left[ \begin{matrix} 
z_1^{(1)} & \dots & z_1^{(m)~} \\ 
\vdots & \ddots & \vdots \\ 
z_{n^{[l]}}^{(1)} & \dots & z_{n^{[l]}}^{(m)~} 
\end{matrix} \right]  
~ \in \mathbb{R}^{n^{[l]} \times m}
\end{alignat}
The associated hidden layer derivatives w.r.t. inputs are stored as: 
\begin{align}
    {\bf A}^{\prime[l]} 
   &=
   \left[
   \begin{matrix}
   {\left[
   \begin{matrix}
   \dfrac{\partial a_1}{\partial x_1}^{(1)} & \dots & \dfrac{\partial a_1}{\partial x_1}^{(m)}  \\
   \vdots & \ddots & \vdots \\
   \dfrac{\partial a_1}{\partial x_{n_x}}^{(1)} & \dots & \dfrac{\partial a_1}{\partial x_{n_x}}^{(m)}\\
   \end{matrix}
   \right]}
   \\ 
   \vdots 
   \\
   {\left[
   \begin{matrix}
   \dfrac{\partial a_{n^{[l]}}}{\partial x_1}^{(1)} & \dots & \dfrac{\partial a_{n^{[l]}}}{\partial x_1}^{(m)}  \\
   \vdots & \ddots & \vdots \\
   \dfrac{\partial a_{n^{[l]}}}{\partial x_{n_x}}^{(1)} & \dots & \dfrac{\partial a_{n^{[l]}}}{\partial x_{n_x}}^{(m)}\\
   \end{matrix}
   \right]}
   \end{matrix}
   \right]
   \in
   \mathbb{R}^{n^{[l]} \times n_x \times m}
   \\
    {\bf Z}^{\prime[l]} 
   &=
   \left[
   \begin{matrix}
   {\left[
   \begin{matrix}
   \dfrac{\partial z_1}{\partial x_1}^{(1)} & \dots & \dfrac{\partial z_1}{\partial x_1}^{(m)}  \\
   \vdots & \ddots & \vdots \\
   \dfrac{\partial z_1}{\partial x_{n_x}}^{(1)} & \dots & \dfrac{\partial z_1}{\partial x_{n_x}}^{(m)}\\
   \end{matrix}
   \right]}
   \\
   \vdots 
   \\ 
   {\left[
   \begin{matrix}
   \dfrac{\partial z_{n^{[l]}}}{\partial x_1}^{(1)} & \dots & \dfrac{\partial z_{n^{[l]}}}{\partial x_1}^{(m)}  \\
   \vdots & \ddots & \vdots \\
   \dfrac{\partial z_{n^{[l]}}}{\partial x_{n_x}}^{(1)} & \dots & \dfrac{\partial z_{n^{[l]}}}{\partial x_{n_x}}^{(m)}\\
   \end{matrix}
   \right]}
   \end{matrix}
   \right]
   \in
   \mathbb{R}^{n^{[l]} \times n_x \times m}
\end{align}
It follows that hidden layer derivatives w.r.t. a specific input $x_j$ are formatted as: 
\begin{alignat}{3}
{\bf A}_j^{\prime[l]} &= \left[ \begin{matrix} 
\dfrac{\partial a_{1}}{\partial x_j}^{(1)} & \dots & \dfrac{\partial a_{1}}{\partial x_j}^{(m)} \\ 
\vdots & \ddots & \vdots \\ 
\dfrac{\partial a_{n^{[l]}}}{\partial x_j}^{(1)} & \dots & \dfrac{\partial a_{n^{[l]}}}{\partial x_j}^{(m)} \\ 
\end{matrix} \right]  
~ \in \mathbb{R}^{n^{[l]} \times m}
\\
{\bf Z}_j^{\prime[l]} &= \left[ \begin{matrix} 
\dfrac{\partial z_{1}}{\partial x_j}^{(1)} & \dots & \dfrac{\partial z_{1}}{\partial x_j}^{(m)} \\ 
\vdots & \ddots & \vdots \\ 
\dfrac{\partial z_{n^{[l]}}}{\partial x_j}^{(1)} & \dots & \dfrac{\partial z_{n^{[l]}}}{\partial x_j}^{(m)} \\ 
\end{matrix} \right]  
~ \in \mathbb{R}^{n^{[l]} \times m}
\end{alignat}
Finally, the back propagation derivatives w.r.t. parameters are: 
\begin{subequations}
\begin{align}
\dfrac{\partial \mathcal{J}}{\partial {\bf W}^{[l]}} &= 
\left[ 
\begin{matrix}
\dfrac{\partial \mathcal{J}}{\partial w_{11}} & \dots & \dfrac{\partial \mathcal{J}}{\partial w_{1n^{[l-1]}}} \\
\vdots & \vdots & \vdots \\
\dfrac{\partial \mathcal{J}}{\partial w_{n^{[l]}1}} & \dots &
\dfrac{\partial \mathcal{J}}{\partial w_{n^{[l]}n^{[l-1]}}}
\end{matrix}
\right]
~ \in \mathbb{R}^{n^{[l]} \times n^{[l-1]}}
\\
\dfrac{\partial \mathcal{J}}{\partial \boldsymbol{b}^{[l]}} &= 
\left[ 
\begin{matrix}
\dfrac{\partial \mathcal{J}}{\partial b_{1}} \\
\vdots  \\
\dfrac{\partial \mathcal{J}}{\partial b_{n^{[l]}}} 
\end{matrix}
\right]
~ \in \mathbb{R}^{n^{[l]} \times 1}
\quad \forall \quad l \in (1,L]
\end{align}
\end{subequations}

\subsection{Forward Propagation (Eq.~\ref{forward_prop})}
\begin{subequations}
\begin{align}
{\bf Z}^{[l]} 
&= {\bf W}^{[l]} \odot {\bf A}^{[l-1]} + \boldsymbol{b}^{[l]} 
&&\quad \forall ~ 1 < l \le L 
\\
{\bf Z}^{^\prime[l]} 
&= {\bf W}^{[l]} \odot {\bf A}_j^{^\prime[l-1]} + \boldsymbol{b}^{[l]} 
&&\quad \forall ~ 1 \le j \le n_x 
\end{align}
\end{subequations}
\begin{subequations}
\begin{align}
{\bf A}^{[l]} &= g\left( {\bf Z}^{[l]} \right)  
&&\quad \forall ~ 1 < l \le L 
\\
{\bf A}_j^{^\prime[l]} 
&= g\left( {\bf Z}^{[l]} \right) \odot {\bf Z}_j^{^\prime[l]}
&&\quad \forall ~ 1 \le j \le n_x 
\end{align}
\end{subequations}

\subsection{Parameter Update (Eq.~\ref{GD})}
\begin{subequations}
\begin{align}
{\bf W}^{[l]} &:= {\bf W}^{[l]} - \alpha \frac{\partial \mathcal{J}}{\partial {\bf W}^{[l]}}
\\ 
\boldsymbol{b}^{[l]} &:= \boldsymbol{b}^{[l]} - \alpha \frac{\partial \mathcal{J}}{\partial \boldsymbol{b}^{[l]}}
&&\quad \forall ~ 1 < l \le L
\end{align}
\end{subequations}
\subsection{Backward Propagation (Eqs.~\ref{eq:dParams}--\ref{eq:last-layer},~\ref{eq:dActivations}, \ref{eq:cost-derivatives})}
The loss function partials w.r.t. last layer activations are: 
\begin{subequations}
\begin{align}
    \frac{\partial \mathcal{L}}{\partial {\bf A}^{[L]}} 
    &= \beta \odot \left( 
        {\bf A}^{[L]} - {\bf Y}
    \right) 
    \\
    \frac{\partial \mathcal{L}}{\partial {\bf A}^{\prime[L]}} 
    &= \gamma \odot \left( 
        {\bf A}^{\prime[L]} - {\bf Y^\prime}
    \right) 
\end{align}
\end{subequations}
The loss function partials w.r.t. hidden layer activations are: 
\begin{subequations}
\begin{align}
\frac{\partial \mathcal{L}}{\partial {\bf A}^{[l-1]}} 
    &= {\bf W}^{\top[l]} \cdot \left(
    g^\prime\left({\bf Z}^{[l]}\right) \odot \frac{\partial \mathcal{L}}{\partial {\bf A}^{[l]}}
    \right) 
    \dots  \\
    \dots &+ \sum_{j=1}^{n_x} 
    {\bf W}^{\top[l]}
    \cdot 
    \left(
    g^{\prime\prime}\left({\bf Z}^{[l]}\right) 
    \odot
    {\bf Z}_{j}^{\prime[l]} 
    \odot
    \frac{\partial \mathcal{L}}{\partial {\bf A}_j^{\prime[l]}}
    \right) \nonumber
\\
\frac{\partial \mathcal{L}}{\partial {\bf A_j}^{\prime[l-1]}} 
&= 
    \sum_{j=1}^{n_x} 
    {\bf W}^{\top[l]}
    \cdot 
    \left(
    g^{\prime}\left({\bf Z}^{[l]}\right) 
    \odot
    \frac{\partial \mathcal{L}}{\partial {\bf A}_j^{\prime[l]}}
    \right)
\end{align}
\end{subequations}
The loss function partials w.r.t. neural net parameters are:
\begin{subequations}
\begin{align}
\frac{\partial \mathcal{L}}{\partial {\bf W}^{[l]}}  
&= 
\frac{\partial \mathcal{L}}{\partial {\bf A}^{[l]}} 
\odot 
g^\prime\left( {\bf Z}^{[l]} \right)
\odot 
{\bf A}^{[l-1]}
+
\sum_{j=1}^{n_x} 
\frac{\partial \mathcal{L}}{\partial {\bf A}_j^{\prime[l]}}
\odot
\bigg[
\dots \nonumber \\
\dots &
g^{\prime\prime}\left( {\bf Z}^{[l]}\right) 
\odot
{\bf Z}_{j}^{\prime[l]} 
\odot 
{\bf A}^{\prime[l-1]}
+
g^{\prime}\left( {\bf Z}^{[l]}\right) {\bf A}_j^{\prime[l-1]}
\bigg]
\\
\frac{\partial \mathcal{L}}{\partial {\bf b}_{[l]}}  
&=  
\frac{\partial \mathcal{L}}{\partial {\bf A}^{[l]}} 
\odot 
g^\prime\left( {\bf Z}^{[l]} \right)
+
\sum_{j=1}^{n_x} 
\frac{\partial \mathcal{L}}{\partial {\bf A}_j^{\prime[l]}}
\odot
g^{\prime\prime}\left( {\bf Z}^{[l]}\right) 
\odot
{\bf Z}_{j}^{\prime[l]} 
\end{align}
\end{subequations}
Finally, the cost function partials w.r.t. neural net parameters are:
\begin{subequations}
\begin{align}
\frac{\partial \mathcal{J}}{\partial {\bf W}^{[l]}} 
&= 
\frac{1}{m}
S
\left( 
\frac{\partial \mathcal{L}}{\partial {\bf W}^{[l]}}  
\right)
+
\frac{\lambda}{m} {\bf W}^{[l]}
\\
\frac{\partial \mathcal{J}}{\partial \boldsymbol{b}^{[l]}} 
&= 
\frac{1}{m} 
S
\left(
\frac{\partial \mathcal{L}}{\partial {\bf W}^{[l]}}  
\right)
\end{align}
\end{subequations}
where S is used to denote the vectorized summation along the last axis of an array\footnote{\url{https://numpy.org/doc/stable/reference/generated/numpy.sum.html}}. Any other summation shown above implies that it could not be vectorized and required a loop. This only happened when summing over partials. Hence, the number of loops needed equals the number of inputs $n_x$, which is usually negligible compared to the size of the training data $m$. The latter is the critical dimension to be vectorized. 

\vskip 0.2in
\bibliography{references}

\end{document}